%% file: main.tex
\pdfoutput=1
\documentclass[11pt]{article}
\usepackage{acl_style/acl}
\usepackage{times}
\usepackage{latexsym}
\usepackage[T1]{fontenc}
\usepackage[utf8]{inputenc}
\usepackage{microtype}
\usepackage{inconsolata}
\usepackage{amsmath} 

\usepackage{hhline}
\usepackage{multirow}
\usepackage{booktabs}
\usepackage{colortbl}
\usepackage[normalem]{ulem}
\useunder{\uline}{\ul}{}
\usepackage{graphicx}
\usepackage{subfig}
\usepackage{svg}
\usepackage{stfloats}
\usepackage{amssymb}
\usepackage{multirow}
\usepackage{xcolor}
\usepackage{enumitem}
\usepackage{marvosym}

\title{PairCFR: Enhancing Model Training on Paired Counterfactually Augmented Data through Contrastive Learning}

\author{Xiaoqi Qiu\textsuperscript{1,*}, Yongjie Wang\textsuperscript{2,*}, Xu Guo\textsuperscript{2}, Zhiwei Zeng\textsuperscript{2}, Yue Yu\textsuperscript{2},\\{\bf Yuhong Feng\textsuperscript{1,\dag}, Chunyan Miao\textsuperscript{2,\dag} }   \\
   \textsuperscript{1} Shenzhen University, \textsuperscript{2} Nanyang Technological University\\ 
   \texttt{\textsuperscript{1} qiuxiaoqi2022@email.szu.edu.cn, yuhongf@szu.edu.cn} \\
   \texttt{\textsuperscript{2}\{yongjie.wang,xu.guo,zhiwei.zeng,yue.yu,ascymiao\}@ntu.edu.sg}
   }

\begin{document}
\maketitle
\begingroup
\renewcommand\thefootnote{\fnsymbol{footnote}}
\footnotetext[1]{Equal contribution.}
\footnotetext[2]{Corresponding author.}
\endgroup

\begin{abstract}
Counterfactually Augmented Data (CAD) involves creating new data samples by applying minimal yet sufficient modifications to flip the label of existing data samples to other classes. Training with CAD enhances model robustness against spurious features that happen to correlate with labels by spreading the casual relationships across different classes.
Yet, recent research reveals that training with CAD may lead models to overly focus on modified features while ignoring other important contextual information, inadvertently introducing biases that may impair performance on out-of-distribution (OOD) datasets.
To mitigate this issue, we employ contrastive learning to promote global feature alignment in addition to learning counterfactual clues. We theoretically prove that contrastive loss can encourage models to leverage a broader range of features beyond those modified ones. 
Comprehensive experiments on two human-edited CAD datasets demonstrate that our proposed method outperforms the state-of-the-art on OOD datasets.

\end{abstract}

\input{introduction}
\input{related_work}

\input{methodv2}

\input{experiment}
\input{conclusion}
\input{limitation}
\input{ethicsConsideration}
%
%
%
\bibliography{ref}
\input{appendix}
\end{document}

%% file: introduction.tex
\section{Introduction}

In the field of Natural Language Processing (NLP), a significant body of research \cite{mccoy-etal-2019-right,wang-culotta-2020-identifying,poliak-etal-2018-hypothesis,gururangan-etal-2018-annotation} has raised the concern that deep learning models can overfit spurious correlations, such as dataset-specific artifacts and biases, rather than focusing on the more complex, generalizable task-related features. For example, \citet{gururangan-etal-2018-annotation} and \citet{poliak-etal-2018-hypothesis} demonstrate that classifiers trained exclusively on hypotheses can still achieve decent results on some Natural Language Inference (NLI) datasets, which ideally requires comparing hypotheses with premises to determine the labels. The existence of biases or shortcuts in training datasets can severely degrade the performance of deep learning models on out-of-distribution (OOD) datasets. 

Counterfactually Augmented Data (CAD) has emerged as a promising approach to mitigate this issue by making minimal modifications to existing data samples such that the corresponding labels are switched to other classes \cite{kaushik2020learning,wen-etal-2022-autocad,pryzant-etal-2023-automatic}. This technique aims to establish direct causal relationships for models to learn more effectively and enhance generalization across different datasets \cite{teney2020learning,kaushik2021explaining}.

However, the effectiveness of CAD is not always guaranteed, particularly when both contexts and the modified information should be considered together to make predictions \cite{joshi-he-2022-investigation,huang-etal-2020-counterfactually}. For instance, in sentiment analysis, simply replacing positive adjectives such as ``good'' or ``excellent'' with negative counterparts like ``terrible'' or ``bad'' will potentially risk models to overemphasize these changes and even assign zero weights to the broader unmodified context \cite{joshi-he-2022-investigation}. Consequently, the trained models may fail to understand more nuanced expressions like irony or negation, exemplified by sentences such as ``Is it a good movie ????'' or ``This movie is not that good.''

To solve the above risks of CAD training, an intuitive solution is to increase the diversity of counterfactual samples \cite{joshi-he-2022-investigation,sen2023people}, thereby disentangling the suspicious correlations between edited features and labels. Nonetheless, this kind of method often relies on human knowledge to steer the diversification, bearing high expenditure and time consumption \cite{huang-etal-2020-counterfactually}. Others try to design additional constraints to align the model gradient with the straight line between the counterfactual example and the original input \citep{teney2020learning}, or to minimize the invariant risk \citep{fan2024unlock}, but these attempts fail to exploit the complex effects of augmented feature components.

In this paper, we introduce a simple yet effective learning strategy to mitigate the overfitting problem associated with CAD. Inspired by the recent success of contrastive learning (CL) in feature alignment \cite{gao-etal-2021-simcse,wang-etal-2022-english,liu-etal-2023-intemats,liu-gran-2023} and its strengths in capturing global relationships \cite{park2022self}, we propose to employ a contrastive learning objective to complement the standard cross-entropy (CE) loss. 
While CL compels the model to extract complementary effects among counterfactually augmented data to alleviate the feature degeneration,
CE ensures the induced feature representations are effectively used for classification. Our mathematical proof further corroborates the advantage of combining the two losses in training models on CAD, resulting in enhanced generation capability.

In summary, our contributions are as follows:
\begin{itemize}[noitemsep,topsep=0pt]
    \item We introduce a contrastive learning-based framework, named Pairwisely Counterfactual Learning with Contrastive Regularization (PairCFR), for training models on CAD, which prevents overfitting to minor, non-robust edits, thus enhancing generalization performance.
    \item We provide theoretical proof for understanding the synergistic benefits of combining the CE and CL losses, unravelling their complementary effects in preventing models from relying solely on counterfactual edits for classification.
    
    \item We conduct comprehensive experiments to demonstrate that the models trained under our learning framework achieve superior OOD generalization performance on two human-edited CAD datasets.
\end{itemize}

%% file: related_work.tex
\section{Related work}

\textbf{Counterfactually Augmented Data.} Counterfactual examples (CFEs)  suggest the minimal modifications required in an input instance to elicit a different outcome \cite{wachter2017counterfactual,barocas2020hidden}. 
This property has inspired researchers~\cite{kaushik2020learning,wu2021polyjuice} to adopt CFEs as a meaningful data augmentation in NLP, aiming to mitigate spurious correlations and improve causal learning. Early efforts~\cite{kaushik2020learning,gardner-etal-2020-evaluating} involved creating CAD datasets with manual sentence edits for label reversal. To ease the high cost of manual annotation, subsequent works adopt large language models (LLMs) for cost-effective generation of CAD~\cite{wu2021polyjuice,madaan2021generate,wen-etal-2022-autocad,dixit-etal-2022-core,pryzant-etal-2023-automatic,chen-etal-2023-disco}.
However, findings from various investigations have indicated that training on CAD does not always ensure improved generalization on OOD tasks~\cite{huang-etal-2020-counterfactually,joshi-he-2022-investigation,fan2024unlock}. 
Consequently, our emphasis in this work is not on generating CAD, but rather on the exploration of methodologies to effectively utilize the inherent prior knowledge within CAD. 

\noindent\textbf{Contrastive Learning.} Contrastive learning is initially proposed to learn a better embedding space by clustering similar samples closely while pushing dissimilar ones far apart~\cite{schroff2015facenet,sohn2016improved,oord2018representation, wang2020hypersphere}.
For example, the triplet loss~\cite{schroff2015facenet} minimizes the distance between an anchor point and its positive sample while maximizing the distance from a negative sample. The N-pair loss~\cite{sohn2016improved} maximizes the distance between an anchor point with multiple negative points. Meanwhile, InfoNCE~\cite{oord2018representation} separates positive samples from multiple noise samples with cross-entropy loss. Enhanced by other efficient techniques, e.g., data augmentation~\cite{chen2020simple}, hard negative sampling~\cite{schroff2015facenet}, and memory bank~\cite{wu2018unsupervised}, CL has propelled significant advancements in various domains, under both supervised and unsupervised settings. In this section, we explore the untapped potential of CL to enhance the OOD generalization of models trained on CAD. 

\noindent\textbf{Training with CAD.} The task of effectively training a robust model with CAD has received relatively limited attention. The simple approach is to directly use the cross-entropy loss~\cite{kaushik2020learning,wen-etal-2022-autocad,balashankar-etal-2023-improving}. To better exploit the causal relationship in counterfactual editing, \newcite{teney2020learning} have introduced gradient supervision over pairs of original data and their counterfactual examples, ensuring the model gradient aligns with the straight line between the original and counterfactual points. Meanwhile,
\newcite{fan2024unlock} considers original and counterfactual distribution as two different environments and proposes a dataset-level constraint using invariant risk minimization. Following these works, we introduce a learning framework employing contrastive loss as a regularizer to enhance the generalization of fine-tuned models notably.

%% file: methodv2.tex
\section{Methodology}
\label{sec:method}

\subsection{Motivation}
Recent studies have empirically shown that while perturbed features in CAD are robust and causal \cite{kaushik2020learning}, they may inhibit the model's ability to learn other robust features that remain unperturbed \cite{joshi-he-2022-investigation}. In this section, we mathematically demonstrate that the standard cross-entropy loss, which is commonly used for training models on CAD, can exacerbate this tendency.


Given an instance $\mathbf{x}\in \mathbb{R}^{m\times 1}$, we train a single-layer non-linear function $f_{W}(x)= \sigma(W^T \mathbf{x})$, where $W \in \mathbb{R}^{m\times 1}$ and $\sigma$ is the sigmoid function, to predict the label $y\in\{0, 1\}$. We expand $\mathbf{x}$, whose label $y=1$, as $\mathbf{x}=[x_r, x_c]^{T}$, where $x_r$ denotes the features to be revised (perturbed) and $x_c$ denotes the constant (unperturbed) features.
The counterfactual example of $\mathbf{x}$ can be written as $\mathbf{c} = [c_r, x_c]^{T}$, with label $y=0$. As the sigmoid function is monotone and bounded, the $c_r$ and $x_r$ should have different signed values to ensure that $\mathbf{x}$ and $\mathbf{c}$ are classified differently.
We expand the weights $W=[w_r, w_c]^{T}$ and take it into the function $f_W$ to obtain $f_W(x)=\sigma(w_r x_r + w_c x_c)$ and $f_W(c)=\sigma(w_r c_r + w_c x_c)$. 
The CE loss on the data $\mathbf{x}$ and its counterfactual $\mathbf{c}$ is calculated as
\begin{align}
    \mathcal{L}_{CE}(\mathbf{x}, \mathbf{c}) &= -\mathrm{log}(f_W(\mathbf{x})) \nonumber\\&- \mathrm{log}(1- f_W(\mathbf{c})).
\end{align}

\noindent By minimizing the CE loss, we enforce $f_W(\mathbf{x})$ to approach 1 and $f_W(\mathbf{c})$ to approach 0. Considering that $x_r$ and its counterpart $c_r$ have different signed values, we observe that optimizing $w_r$ can achieve the desired contrasting effect with less effort than optimizing $w_c$. Therefore, the model tends to assign higher weights $w_r$ for revised features and lower weights $w_c$ for constant or unperturbed features. An expanded illustration can be found in the Appendix \ref{sec:sigmoid_function}.
Similar phenomena are observed in both least squares loss \cite{joshi-he-2022-investigation} and Fisher’s Linear Discriminant on CAD \cite{fan2024unlock}. 

The above observations indicate that the CE loss alone can lead the model to focus on learning the revised features in CAD, which necessitates incorporating a regularization that compels the model to consider a broader range of features.

\subsection{The Role of Contrastive Loss}
\label{subsec: cl reg}
Recent research findings have empirically shown that models trained under contrastive loss mainly focus on capturing global relationships \cite{park2022self} compared with negative log-likelihood losses such as masked language modeling. Inspired by this, we propose to employ CL to complement standard CE loss for training models on CAD. In the following, we start from the introduction of CL loss and then mathematically show how CL encourages the model to select a broader range of features beyond the edited ones in the counterfactual data.


%
Given an anchor sample $\mathbf{x}_i$ from a data batch $\mathcal{D}=\{\mathbf{x}_i, y_i\}_{i=1}^N$, $\forall \mathbf{x}_i \in \mathcal{D}$, we have its positive samples in $\mathcal{P}_{i}\!\equiv\!\{\mathbf{x}_p|y_{p}=y_{i},p\neq i\}$ and negative samples in $\mathcal{N}_{i}\!\equiv\!\{\mathbf{x}_n|y_{n}\neq y_{i}, n\neq i\}$, where $\mathcal{N}_{i}$ contains the counterfactual samples $\mathbf{c}$ for every $\mathbf{x}_i$. The contrastive loss for the anchor $\mathbf{x}_i$ is
\begin{equation}
    \mathcal{L}_{CL} \!=\! -\underset{\mathbf{x}_p\in\mathcal{P}_{i}}{\mathbb{E}}\left[\mathrm{log}\frac{e^{s_{ip}/\tau}}{e^{s_{ip}/\tau} + \mathop{\sum}_{\mathbf{x}_n\in\mathcal{N}_i}e^{s_{in}/\tau}}\right],
\end{equation}

\noindent where $s_{xy}=\frac{\mathbf{z}_{x} \cdot \mathbf{z}_{y}}{\parallel \mathbf{z}_x \parallel\parallel \mathbf{z}_y \parallel}$ measures the cosine similarity between the hidden representations of a pair of samples, and $\tau$ is a temperature scaling factor for controlling the extent to which we separate positive and negative pairs \cite{wang2020hypersphere}.

Without loss of generality, we assume $\mathbf{W} \in \mathbb{R}^{m \times d}$ that directly maps the input instance into a $d$-dimensional embedding space, $\mathbf{z}_i = \mathbf{W}^T \mathbf{x}_i$. 
To obtain the gradient of the CL loss coming from negative samples, we have
\begin{align}
 \frac{\partial\mathcal{L}_{CL}}{\partial\mathbf{W}}\bigg|_{s_{in}} &=\frac{\partial \mathcal{L}_{CL} }{\partial s_{in}}\times \frac{\partial s_{in}}{\partial \mathbf{W} }
  \nonumber \\ 
  &=\frac{1}{\tau } P_{in}\times\mathbf{A}_{in}\mathbf{W}.
\end{align}
The full derivation process can be found in the appendix \ref{sec:gradient_analysis}. Here, we have
\begin{align}
    P_{in} \!=\! \underset{\mathbf{x}_p\in\mathcal{P}_{i}}{\mathbb{E}}\!\!\left [\!
    \frac{e^{s_{in}/\tau}}{e^{s_{ip}/\tau}\!+\!\underset{\mathbf{x}_n\in \mathcal{N}_{i}}{\sum}e^{s_{in}/\tau}\!}\right ]\! ,
\end{align}
which indicates the probability of $\mathbf{x}_i$ being recognized as $\mathbf{x}_n$. $\mathbf{A}_{in}\!=\!\mathbf{x}_{i}\mathbf{x}_{n}^{T}+\mathbf{x}_{n}\mathbf{x}_{i}^{T}\in \mathbb{R}^{m\times m}$ is a symmetric matrix derived from the outer product of $\mathbf{x}_{i}$ and $\mathbf{x}_{n}$. Each element of $\mathbf{A}_{i,n}$ indicates the digit-level dot product between the features of $\mathbf{x}_i$ and $\mathbf{x}_n$, which provides a full view of the entire feature space when comparing a pair of samples. A higher value leads to a larger gradient update and the weights $\mathbf{W}$ are optimized by considering the whole feature sets. 

The above analysis implies that the CL loss has the capability of capturing global features
beyond those being edited.  
When learning on CAD under CL, we pair each instance $\mathbf{x}$ with its CFE, $\mathbf{c}$, to compel the model to disparate $\mathbf{x}$ from all negative samples, including its counterfactual example $\mathbf{c}$:
\begin{align}
\underset{\mathbf{x}_n\in \mathcal{N}_{i}}{\sum}e^{s_{in}/\tau} \!=\! e^{s_{ic}/\tau} \!+\! \underset{\mathbf{x}_n\in \mathcal{N}_{i}\backslash \mathbf{c}}{\sum}e^{s_{in}/\tau},
\end{align}
where the non-bold $c$ is the index of CFE. Let us revisit the toy example with $\mathbf{x} = [x_r, x_c]^{T}$ and $\mathbf{c} = [c_r, x_c]^{T}$. Although minimizing the similarity between $\mathbf{x}$ and $\mathbf{c}$ encourages the model to focus on features $x_r$, the other negative samples in the batch, e.g., $\mathbf{x}' = [x_r', x_c']^{T}$, will enforce the model to use both $w_r$ and $w_c$ to compare the difference. Hence, the existence of real negative samples could help the model capture the relationships between $x_r$ and its context $x_c$. 

As all $s_{in}$ equally contribute to updating the model weights, the number of non-CFE negatives moderates the learning from local CAD and global patterns. A smaller batch size will manifest the influence of edited features, whereas a larger batch size may dilute the local differences in CAD, as discussed in the experiments \ref{exp:batch_size}.

\begin{figure}[bt]
    \centering
    \includegraphics[scale=0.58]{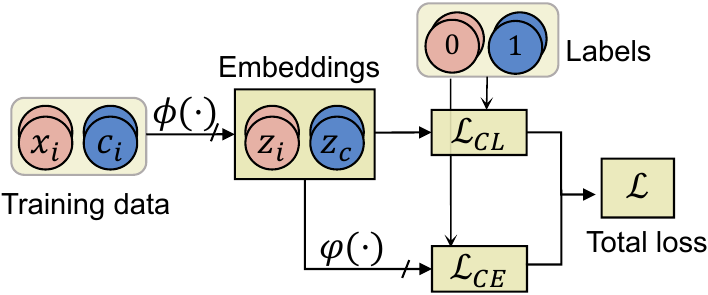}
    \caption{The overall learning framework.}
    \label{fig:systemdraft}
\end{figure}

\subsection{Overall Learning Framework}
Next, we introduce our proposed learning framework, Pairwisely Counterfactual Learning with Contrastive Loss Regularization, named PairCFR for short. As shown in Figure \ref{fig:systemdraft}, a model $f$ can be decomposed into two modules, $\phi(\cdot)$ and $\varphi(\cdot)$, i.e., $f = \varphi \circ \phi$, where $\phi(\cdot)$ encodes the input sentence into a hidden embedding, and $\varphi(\cdot)$ maps $\phi(\mathbf{x})$ for classification. For transformer-based models, 
we instantiate $\phi(\mathbf{x})$ using the [CLS] representation, denoted as $\mathbf{z}$. 
We explicitly pair the original sentences $\mathbf{x}$ and their CFEs, $\mathbf{c}$, in the same batch to provide additional training signals indicative of the underlying causal relationships. 

The standard cross-entropy loss is computed on the logits vector projected from $\varphi(z)$.
Optimizing CE loss enforces $\varphi(\cdot)$ to identify a small set of features from $\mathbf{z}$ and assign them higher weights to quickly reach a local minimum while optimizing CL loss compels $\phi(\cdot)$ to consider the entire feature space of $\mathbf{z}$ to meet the distance constraints.
Overall, we combine the two losses as follows.
\begin{equation}
    \mathcal{L} = \lambda \mathcal{L}_{CL} + (1-\lambda) \mathcal{L}_{CE},
\end{equation}
where $\lambda$ is the trade-off factor to balance the two losses. To compute CL on a batch, we sample positive pairs that have the same label while all the negative samples including the CFE of the anchor sample are considered.

%% file: experiment.tex
\section{Experimental Setup }
In the following, we introduce experimental settings, which include benchmark tasks, evaluation metrics, competitive baselines and implementation details. Our code is released on GitHub \footnote{https://github.com/Siki-cloud/PairCFR.git}. 

\subsection{Benchmark Tasks \& Evaluations}

We evaluate our learning framework on two NLP tasks, sentiment analysis (SA) and natural language inference (NLI). We use two human-edited CAD datasets \cite{kaushik2020learning}, which ensures good-quality counterfactual data \cite{sen2023people}, to train all the models.
The IMDb augmented dataset contains 4880 data samples with an original to CFE ratio of 1:1. The SNLI dataset contains 11330 data samples with an original to CFE ratio of 1:4. 
The statistics of human-revised CAD are reported in Appendix \ref{apd:train data}. 


To eliminate the random effect, we train each model for multiple runs ($10$ runs for SA and $7$ runs for NLI) using different random seeds. We report the average test \textbf{accuracy}, \textbf{standard deviation} for both in-domain (ID) datasets and several out-of-domain (OOD) datasets. We also conduct significance tests by calculating \textbf{p-value}, to ensure that the observed improvements are not due to randomness. The details of ID and OOD datasets used for evaluation are described in Appendix \ref{apd:id and ood tests}.


\subsection{Implementation Details}
We finetune the BERT base \cite{devlin2019bert}, RoBERTa base \cite{liu2019roberta}, Sentences-BERT (SBERT, multi-qa-distilbert-cos) \cite{reimers2019sbert} and T5 base \cite{2020t5} models with the original or CAD datasets on  HuggingFace platform \cite{wolf-etal-2020-transformers}. Volumes of model parameters are listed in Table \ref{tab:modelsize} in Appendix \ref{apd:hyperparameter values}. Following the common practices of transformers \cite{devlin2019bert}, we take the embedding of the ``[CLS]'' token as sentence representation and finetune the whole model. We set the maximum token length to 350 for SA and 64 for NLI. 

We follow the original dataset splits described in \cite{kaushik2020learning}, where the train, validation, and test sets are divided in a ratio of 7:1:2, with all classes balanced across each set. Subsequently, we finetune all models up to 20 epochs with the AdamW optimizer, coupled with a linear learning rate scheduler with a warmup ratio as 0.05. The best learning rate is manually tuned from $\{1e^{-4},1e^{-5},3e^{-5},5e^{-5},5e^{-6},1e^{-6}\}$. We apply the early stopping strategy with a patience of $5$ and the best model is selected according to the lowest validation loss. To determine the trade-off factor $\lambda$ and temperature $\tau$, we conducted a grid search in the range $[0, 1]$ with a step size of 0.1.  We also conducted experiments to evaluate our PairCFR in few shot setting where the learning rate and batch size were tuned accordingly. The hyperparameters for full data finetuning and few shot setting are shown in Table \ref{tab:few shot parameter}, Table \ref{tab:optimal parameters} respectively, in Appendix \ref{apd:hyperparameter values}.

\subsection{Baselines}
We compare our method PairCFR with the following baselines. For a fair comparison, we employ other forms of augmentation or increase the sampling number for the first three baselines without counterfactual augmentation, to ensure all approaches have the same number of training data. 

\textbf{Vanilla} \cite{devlin2019bert}. This method refers to a general model fine-tuning with original sentences. We include this baseline to verify the improvement of our method result from both the introduction of CAD and the novel learning framework. 

\textbf{BTSCL} \cite{gunel2021supervised}. This approach employs the supervised contrastive loss \cite{khosla2020supervised} into the model training where augmented positive samples are obtained through back-translating a given sentence \cite{ng2020translation}. 

\textbf{CouCL} \cite{wang2022counterexample}. As counterexamples (CEs) are rare in a mini-batch, CouCL samples counterexamples from the original training set, where an example with lower confidence corresponds to a higher likelihood of being selected. Subsequently, it adopts the self-supervised contrastive loss to push representations of positive CEs and negative CEs far apart. 

The following approaches study how to train a robust model with annotated CAD:

\textbf{HCAD} \cite{kaushik2020learning}. It collects two human-edited CAD datasets and fine-tunes a pretrained model on CAD with the cross-entropy loss.

\textbf{CFGSL} \cite{teney2020learning}. As domain priors in CAD may be lost due to random shuffling in preprocessing \cite{kaushik2020learning}, CFGSL pairs original data and its counterfactual example in the same batch and introduces a gradient supervision loss (GSL) alongside the cross-entropy loss. The GSL enforces the model gradient to align with the straight line from the original point to CFE.

\textbf{ECF}  \cite{fan2024unlock}. It introduces two additional losses to mine the causal structures of CAD. The first loss extracts the dataset-level invariance through Invariant Risk Minimization (IRM) while the second loss is applied to pairs of original sentences and CFEs, preventing the model from relying on correlated features. 




\begin{table*}[]
\caption{Average performance of various fine-tuned models on ID and OOD test sets. $\overline{Acc}$ denotes the average of all the OOD performance. The best results are bolded.}
\label{tab:full}
\tabcolsep=0.11cm
\scalebox{0.63}{
\begin{tabular}{@{}lccccccccccccc@{}}
\toprule
\multicolumn{1}{c}{\multirow{3}{*}{Methods}} & \multicolumn{6}{c}{Sentiment Analysis} & \multicolumn{7}{c}{Natural Language Inference} \\ \cmidrule(l){2-14} 
\multicolumn{1}{c}{} & \multicolumn{1}{c|}{In-Domain} & \multicolumn{5}{c||}{Out-of-Dimain} & \multicolumn{1}{c|}{In-Domain} & \multicolumn{6}{c}{Out-of-Dimain} \\ \cmidrule(l){2-14} 
\multicolumn{1}{c}{} & \multicolumn{1}{c|}{IMDb} & Amazon & Yelp & Twitter & SST-2 & \multicolumn{1}{c||}{$\overline{Acc}$} & \multicolumn{1}{c|}{SNLI} & MNLI-m & MNLI-mm & Negation & Spelling-e & Word-o & \multicolumn{1}{l}{$\overline{Acc}$} \\ \midrule
\multicolumn{14}{c}{BERT-base-uncased} \\ \midrule
Vanilla & \multicolumn{1}{c|}{90.15±1.66} & 86.38±0.39 & 91.03±0.83 & 81.66±0.27 & 82.59±1.00 & \multicolumn{1}{c||}{85.42} & \multicolumn{1}{c|}{78.85±0.44} & 57.43±0.92 & 59.36±0.80 & 40.96±4.32 & 53.56±1.54 & 50.75±6.65 & 52.41 \\
BTSCL & \multicolumn{1}{c|}{\textbf{90.43±1.47}} & 85.45±0.71 & \textbf{91.97±0.31} & 81.79±1.28 & 83.80±1.17 & \multicolumn{1}{c||}{85.75} & \multicolumn{1}{c|}{\textbf{79.02±0.49}} & 57.28±1.30 & 59.10±1.42 & \textbf{43.10±3.65} & 53.51±1.74 & 49.20±4.51 & 52.44 \\
CouCL & \multicolumn{1}{c|}{85.67±1.13} & 86.75±0.22 & 89.53±0.55 & 84.41±0.23 & 85.01±0.43 & \multicolumn{1}{c||}{86.43} & \multicolumn{1}{c|}{71.90±0.95} & 51.99±1.75 & 52.20±1.86 & 38.70±4.69 & 49.82±2.01 & 44.03±4.02 & 47.35 \\
HCAD & \multicolumn{1}{c|}{88.16±2.70} & 86.40±0.77 & 89.94±0.99 & 83.29±2.71 & 85.74±1.04 & \multicolumn{1}{c||}{86.34} & \multicolumn{1}{c|}{73.49±1.37} & 58.53±1.59 & 60.77±1.46 & 35.43±3.06 & 54.01±2.70 & 54.72±3.29 & 52.69 \\
CFGSL & \multicolumn{1}{c|}{88.51±3.29} & 85.52±1.05 & 89.58±1.83 & 84.56±1.53 & 86.77±0.79 & \multicolumn{1}{c||}{86.61} & \multicolumn{1}{c|}{77.16±0.41} & 60.11±1.07 & 62.25±0.66 & 33.81±1.89 & 56.37±0.74 & 58.45±0.97 & 54.20 \\
ECF & \multicolumn{1}{c|}{87.71±0.29} & 86.43±0.10 & 89.30±0.16 & 83.05±0.69 & 86.23±0.18 & \multicolumn{1}{c||}{86.25} & \multicolumn{1}{c|}{73.23±1.52} & 58.95±0.15 & 61.19±1.34 & 42.40±1.07 & 54.15±0.53 & 57.10±0.92 & 54.76 \\
Ours & \multicolumn{1}{c|}{89.63±1.36} & \textbf{86.79±0.72} & 91.78±0.44 & \textbf{85.27±0.39} & \textbf{86.81±0.97} & \multicolumn{1}{c||}{\textbf{87.66}} & \multicolumn{1}{c|}{75.38±0.21} & \textbf{60.46±0.38} & \textbf{62.27±0.39} & 39.21±3.61 & \textbf{56.84±0.54} & \textbf{59.16±0.88} & \textbf{55.59} \\ \midrule
\multicolumn{14}{c}{RoBERTa-base} \\ \midrule
Vanilla & \multicolumn{1}{c|}{92.68±1.15} & 87.08±1.39 & 94.00±0.77 & 81.43±2.82 & 86.04±2.76 & \multicolumn{1}{c||}{87.14} & \multicolumn{1}{c|}{85.16±0.39} & 70.35±1.29 & 71.25±1.59 & 52.47±5.55 & 67.36±1.36 & 61.82±4.54 & 64.65 \\
BTSCL & \multicolumn{1}{c|}{\textbf{93.09±0.61}} & 89.46±0.21 & \textbf{94.74±0.36} & 85.72±1.22 & 87.16±0.87 & \multicolumn{1}{c||}{89.27} & \multicolumn{1}{c|}{\textbf{85.72±0.44}} & 70.83±1.38 & 72.10±1.32 & \textbf{56.89±3.78} & 67.61±1.32 & 62.22±3.55 & 65.93 \\
CouCL & \multicolumn{1}{c|}{91.22±0.83} & 89.48±0.19 & 93.04±0.58 & 87.40±0.77 & 88.07±0.66 & \multicolumn{1}{c||}{89.50} & \multicolumn{1}{c|}{82.37±0.52} & 70.86±1.32 & 71.38±1.23 & 51.83±2.71 & 68.08±1.23 & 64.68±1.82 & 65.37 \\
HCAD & \multicolumn{1}{c|}{90.12±1.74} & 88.50±0.57 & 92.18±0.94 & 83.43±1.75 & 86.48±0.98 & \multicolumn{1}{c||}{87.65} & \multicolumn{1}{c|}{80.91±0.69} & 70.35±1.08 & 70.77±0.76 & 45.79±4.16 & 67.37±1.28 & 64.83±1.47 & 63.82 \\
CFGSL & \multicolumn{1}{c|}{90.69±0.92} & 88.32±0.41 & 93.48±0.48 & 83.90±1.78 & 86.89±0.80 & \multicolumn{1}{c||}{88.15} & \multicolumn{1}{c|}{82.45±0.35} & 71.59±0.90 & 71.25±1.06 & 51.40±1.47 & 68.86±1.07 & 62.22±1.99 & 65.06 \\
ECF & \multicolumn{1}{c|}{91.05±0.44} & 88.56±0.32 & 93.79±0.19 & 85.82±0.43 & 87.84±0.59 & \multicolumn{1}{c||}{89.00} & \multicolumn{1}{c|}{81.88±0.17} & 70.45±1.03 & 71.18±0.93 & 51.70±2.38 & 66.60±0.94 & 63.76±1.98 & 64.74 \\
Ours & \multicolumn{1}{c|}{91.74±0.88} & \textbf{89.60±0.26} & 93.35±0.34 & \textbf{87.90±0.45} & \textbf{88.61±0.41} & \multicolumn{1}{c||}{\textbf{89.87}} & \multicolumn{1}{c|}{82.13±0.51} & \textbf{71.80±0.53} & \textbf{72.12±0.79} & 55.19±1.97 & \textbf{68.88±0.36} & \textbf{65.91±1.35} & \textbf{66.78} \\ \midrule
\multicolumn{14}{c}{SBERT-multi-qa-distilbert-cos} \\ \midrule
Vanilla & \multicolumn{1}{c|}{87.61±1.86} & 80.65±0.67 & 89.74±0.77 & 83.95±1.12 & 82.01±1.59 & \multicolumn{1}{c||}{84.09} & \multicolumn{1}{c|}{76.96±0.53} & 53.90±2.03 & 55.90±2.22 & 45.20±4.18 & 51.23±2.72 & 48.27±5.00 & 50.90 \\
BTSCL & \multicolumn{1}{c|}{\textbf{88.84±2.41}} & 81.21±0.76 & \textbf{90.49±0.37} & 84.20±0.61 & 83.62±0.64 & \multicolumn{1}{c||}{84.88} & \multicolumn{1}{c|}{\textbf{77.16±0.38}} & 54.42±1.31 & 56.14±1.36 & \textbf{45.40±2.78} & 52.44±1.83 & 49.80±2.63 & 51.64 \\
CouCL & \multicolumn{1}{c|}{87.96±0.67} & 83.92±0.13 & 89.15±0.18 & 85.40±0.31 & 83.48±0.37 & \multicolumn{1}{c||}{85.49} & \multicolumn{1}{c|}{70.61±1.54} & 55.29±1.45 & 57.90±1.81 & 35.86±1.87 & 52.01±2.26 & 54.89±1.91 & 51.19 \\
HCAD & \multicolumn{1}{c|}{86.09±1.74} & 83.94±0.39 & 87.87±0.66 & 85.91±0.66 & 82.83±0.90 & \multicolumn{1}{c||}{85.14} & \multicolumn{1}{c|}{71.64±1.04} & 55.93±1.61 & 58.70±1.96 & 35.05±1.22 & 53.33±1.06 & 54.86±2.08 & 51.57 \\
CFGSL & \multicolumn{1}{c|}{86.05±1.07} & 82.71±0.73 & 87.59±0.75 & 83.36±0.55 & 83.70±0.49 & \multicolumn{1}{c||}{84.34} & \multicolumn{1}{c|}{70.72±1.06} & 55.84±0.88 & 58.52±1.15 & 36.07±3.38 & 52.60±1.27 & 55.57±1.68 & 51.72 \\
ECF & \multicolumn{1}{c|}{87.83±0.46} & 84.51±0.34 & 88.44±0.20 & 84.60±0.70 & 84.27±0.56 & \multicolumn{1}{c||}{85.46} & \multicolumn{1}{c|}{64.55±1.23} & 49.95±1.84 & 51.49±1.82 & 38.59±2.32 & 48.31±1.67 & 49.55±2.27 & 47.58 \\
Ours & \multicolumn{1}{c|}{87.28±0.75} & \textbf{84.58±0.22} & 88.52±0.30 & \textbf{86.32±0.35} & \textbf{84.31±0.78} & \multicolumn{1}{c||}{\textbf{85.93}} & \multicolumn{1}{c|}{71.48±0.40} & \textbf{57.19±0.84} & \textbf{60.76±0.46} & 37.27±2.35 & \textbf{54.36±0.67} & \textbf{56.78±1.24} & \textbf{53.27} \\ \midrule
\multicolumn{14}{c}{T5-base} \\ \midrule
Vanilla & \multicolumn{1}{c|}{92.15±1.49} & 88.24±0.85 & 94.44±0.67 & 83.40±1.38 & 86.17±2.60 & \multicolumn{1}{c||}{88.06} & \multicolumn{1}{c|}{83.28±0.57} & 62.62±2.59 & 65.18±2.10 & 41.00±2.46 & 58.76±2.61 & 48.30±3.27 & 55.17 \\
BTSCL & \multicolumn{1}{c|}{\textbf{92.78±1.08}} & 88.50±0.81 & \textbf{94.89±0.42} & 83.37±1.09 & 87.17±1.07 & \multicolumn{1}{c||}{88.48} & \multicolumn{1}{c|}{\textbf{83.66±0.46}} & 64.01±2.57 & 66.47±2.24 & 42.16±2.90 & 60.01±3.43 & 50.16±5.69 & 56.56 \\
CouCL & \multicolumn{1}{c|}{91.74±0.88} & 88.91±0.47 & 93.35±0.34 & 87.03±0.70 & 88.61±0.41 & \multicolumn{1}{c||}{89.48} & \multicolumn{1}{c|}{79.81±0.54} & 70.19±0.58 & 71.84±0.76 & 39.82±3.23 & 66.35±0.68 & 64.29±1.58 & 62.50 \\
HCAD & \multicolumn{1}{c|}{90.09±1.95} & 88.72±0.85 & 92.60±0.87 & 85.63±1.15 & 85.54±1.28 & \multicolumn{1}{c||}{88.12} & \multicolumn{1}{c|}{80.09±0.73} & 70.19±0.72 & 71.60±0.83 & 45.05±3.94 & 66.57±0.73 & 65.30±1.51 & 63.74 \\
CFGSL & \multicolumn{1}{c|}{89.48±5.17} & 88.27±1.05 & 92.77±1.45 & 81.56±2.49 & 82.11±2.50 & \multicolumn{1}{c||}{86.18} & \multicolumn{1}{c|}{80.71±0.64} & 69.08±0.97 & 69.85±1.12 & 45.59±3.74 & 65.58±1.18 & 65.80±1.55 & 63.18 \\
ECF & \multicolumn{1}{c|}{90.85±0.37} & 89.27±0.25 & 92.65±0.44 & 87.66±0.26 & 88.57±0.54 & \multicolumn{1}{c||}{89.54} & \multicolumn{1}{c|}{78.93±0.51} & 69.57±1.14 & 70.30±1.45 & 46.14±3.12 & 64.19±1.08 & 65.79±1.71 & 63.20 \\
Ours & \multicolumn{1}{c|}{91.47±0.89} & \textbf{89.18±0.21} & 93.45±0.63 & \textbf{87.90±0.45} & \textbf{88.64±1.04} & \multicolumn{1}{c||}{\textbf{89.79}} & \multicolumn{1}{c|}{80.87±0.77} & \textbf{71.38±0.13} & \textbf{72.46±0.57} & \textbf{46.31±0.50} & \textbf{67.37±0.12} & \textbf{67.39±0.33} & \textbf{64.98} \\ \bottomrule
\end{tabular}}
\end{table*}


\begin{table*}[]
\caption{Ablation study for the pairing strategy and the CL loss on various transformer-based models. $\overline{Acc}$ denotes the average of all the OOD performance. The best results are bolded.}
\label{tab:abalation}
\tabcolsep=0.11cm
\scalebox{0.6}{
\begin{tabular}{@{}lllllllcllllllc@{}}
\toprule
\multicolumn{2}{c}{} & \multicolumn{6}{c}{Sentiment Analysis} & \multicolumn{7}{c}{Natural Language Inference} \\ \midrule
\multicolumn{2}{c}{Variants} & \multicolumn{1}{c|}{In-Domain} & \multicolumn{5}{c||}{Out-of-Dimain} & \multicolumn{1}{c|}{In-Domain} & \multicolumn{6}{c}{Out-of-Dimain} \\ \midrule
\#Train & \multicolumn{1}{c}{Loss} & \multicolumn{1}{c|}{IMDb} & \multicolumn{1}{c}{Amazon} & \multicolumn{1}{c}{Yelp} & \multicolumn{1}{c}{Twitter} & \multicolumn{1}{c}{SST-2} & \multicolumn{1}{l||}{$\overline{Acc}$} & \multicolumn{1}{c|}{SNLI} & \multicolumn{1}{c}{MNLI-m} & \multicolumn{1}{c}{MNLI-mm} & \multicolumn{1}{c}{Negation} & \multicolumn{1}{c}{Spelling-e} & \multicolumn{1}{c}{Word-o} & \multicolumn{1}{l}{$\overline{Acc}$} \\ \midrule
\multicolumn{15}{c}{BERT-base-uncased} \\ \midrule
ShuffCAD & CE & \multicolumn{1}{l|}{88.16±2.70} & 86.40±0.77 & 89.94±0.99 & 83.29±2.71 & 85.74±1.04 & \multicolumn{1}{c||}{86.34} & \multicolumn{1}{l|}{73.49±1.37} & 58.53±1.59 & 60.77±1.46 & 35.43±3.06 & 54.01±2.70 & 54.72±3.29 & 52.69 \\
PairCAD & CE & \multicolumn{1}{l|}{88.23±3.11} & 86.56±0.34 & 89.97±1.85 & 84.15±1.20 & 85.84±0.85 & \multicolumn{1}{c||}{86.62} & \multicolumn{1}{l|}{74.27±0.72} & 59.13±0.65 & 60.85±0.88 & 36.10±1.92 & 56.14±1.34 & 55.40±2.83 & 53.52 \\
ShuffCAD & CE+CL & \multicolumn{1}{l|}{89.18±1.33} & 86.77±0.65 & 91.45±0.53 & 84.14±1.82 & 86.26±0.99 & \multicolumn{1}{c||}{87.15} & \multicolumn{1}{l|}{73.77±1.11} & 59.39±0.64 & 61.85±0.86 & 36.80±4.04 & 55.62±0.87 & 57.09±2.45 & 54.15 \\
PairCAD & CE+CL & \multicolumn{1}{l|}{\textbf{89.63±1.36}} & \textbf{86.79±0.72} & \textbf{91.78±0.44} & \textbf{85.27±0.39} & \textbf{86.81±0.97} & \multicolumn{1}{c||}{\textbf{87.66}} & \multicolumn{1}{l|}{\textbf{75.38±0.21}} & \textbf{60.46±0.38} & \textbf{62.27±0.39} & \textbf{39.21±3.61} & \textbf{56.84±0.54} & \textbf{59.16±0.88} & \textbf{55.59} \\ \midrule
\multicolumn{15}{c}{RoBERTa-base} \\ \midrule
ShuffCAD & CE & \multicolumn{1}{l|}{90.12±1.74} & 88.50±0.57 & 92.18±0.94 & 83.43±1.75 & 86.48±0.98 & \multicolumn{1}{c||}{87.67} & \multicolumn{1}{l|}{80.91±0.69} & 70.35±1.08 & 70.77±0.76 & 45.79±4.16 & 67.37±1.28 & 64.83±1.47 & 63.82 \\
PairCAD & CE & \multicolumn{1}{l|}{90.95±0.84} & 88.77±0.74 & 92.77±0.95 & 83.45±2.53 & 86.37±1.06 & \multicolumn{1}{c||}{87.84} & \multicolumn{1}{l|}{81.69±0.90} & 70.77±0.49 & 71.33±0.45 & 54.38±1.67 & 67.90±0.63 & 65.43±0.99 & 65.96 \\
ShuffCAD & CE+CL & \multicolumn{1}{l|}{91.42±1.01} & 89.44±0.27 & 92.91±0.64 & 86.67±1.05 & 87.25±0.68 & \multicolumn{1}{c||}{89.07} & \multicolumn{1}{l|}{81.95±0.39} & 71.16±0.60 & 71.79±0.79 & 51.43±2.91 & 68.20±0.57 & 64.12±1.03 & 65.34 \\
PairCAD & CE+CL & \multicolumn{1}{l|}{\textbf{91.74±0.88}} & \textbf{89.60±0.26} & \textbf{93.35±0.34} & \textbf{87.90±0.45} & \textbf{88.61±0.41} & \multicolumn{1}{c||}{\textbf{89.61}} & \multicolumn{1}{l|}{\textbf{82.13±0.51}} & \textbf{71.80±0.53} & \textbf{72.12±0.79} & \textbf{55.19±1.97} & \textbf{68.88±0.36} & \textbf{65.91±1.35} & \textbf{66.78} \\ \midrule
\multicolumn{15}{c}{SBERT-multi-qa-distilbert-cos} \\ \midrule
ShuffCAD & CE & \multicolumn{1}{l|}{86.09±1.74} & 83.94±0.39 & 87.87±0.66 & 85.91±0.66 & 82.83±0.90 & \multicolumn{1}{c||}{85.13} & \multicolumn{1}{l|}{\textbf{71.64±1.04}} & 55.93±1.61 & 58.70±1.96 & 35.05±1.22 & 53.33±1.06 & 54.86±2.08 & 51.57 \\
PairCAD & CE & \multicolumn{1}{l|}{86.78±1.41} & 83.55±0.39 & 88.51±0.77 & 85.95±0.40 & 83.20±0.63 & \multicolumn{1}{c||}{85.30} & \multicolumn{1}{l|}{70.90±1.02} & 56.50±0.58 & 59.03±0.57 & 35.89±1.98 & 53.03±1.17 & 55.04±1.03 & 51.89 \\
ShuffCAD & CE+CL & \multicolumn{1}{l|}{\textbf{87.68±1.05}} & 84.23±0.37 & 88.66±0.77 & 85.45±0.28 & 83.60±0.38 & \multicolumn{1}{c||}{85.48} & \multicolumn{1}{l|}{71.38±0.62} & 57.08±0.53 & 60.01±0.35 & 35.11±1.64 & 54.15±0.53 & 55.59±1.89 & 52.39 \\
PairCAD & CE+CL & \multicolumn{1}{l|}{87.28±0.22} & \textbf{84.58±0.22} & \textbf{88.52±0.30} & \textbf{86.32±0.35} & \textbf{84.31±0.7} & \multicolumn{1}{c||}{\textbf{85.93}} & \multicolumn{1}{l|}{71.48±0.40} & \textbf{57.19±0.84} & \textbf{60.76±0.46} & \textbf{37.27±2.35} & \textbf{54.36±0.67} & \textbf{56.78±1.24} & \textbf{53.27} \\ \midrule
\multicolumn{15}{c}{T5-base} \\ \midrule
ShuffCAD & CE & \multicolumn{1}{l|}{90.09±1.95} & 88.72±0.85 & 92.60±0.87 & 85.63±1.15 & 85.54±1.28 & \multicolumn{1}{c||}{88.12} & \multicolumn{1}{l|}{80.09±0.73} & 70.19±0.72 & 71.60±0.83 & 45.05±3.94 & 66.57±0.73 & 65.30±1.51 & 63.85 \\
PairCAD & CE & \multicolumn{1}{l|}{90.03±1.35} & 89.02±0.41 & 92.76±0.99 & 86.46±1.00 & 86.59±1.37 & \multicolumn{1}{c||}{88.71} & \multicolumn{1}{l|}{79.55±0.66} & 68.86±0.52 & 70.75±0.77 & 45.18±3.49 & 65.56±0.67 & 65.64±1.50 & 62.83 \\
ShuffCAD & CE+CL & \multicolumn{1}{l|}{90.38±1.80} & 89.03±0.46 & 93.06±1.29 & 85.75±0.96 & 87.24±2.12 & \multicolumn{1}{c||}{88.76} & \multicolumn{1}{l|}{80.21±0.10} & 70.43±0.11 & 71.78±0.37 & 45.41±2.08 & 66.59±0.56 & 66.28±0.93 & 64.09 \\
PairCAD & CE+CL & \multicolumn{1}{l|}{\textbf{91.47±0.89}} & \textbf{89.18±0.21} & \textbf{93.45±0.63} & \textbf{87.03±0.70} & \textbf{88.64±1.04} & \multicolumn{1}{c||}{\textbf{89.79}} & \multicolumn{1}{l|}{\textbf{80.87±0.77}} & \textbf{71.38±0.13} & \textbf{72.46±0.57} & \textbf{46.31±0.50} & \textbf{67.37±0.12} & \textbf{67.39±0.33} & \textbf{64.98} \\ \bottomrule
\end{tabular}}
\end{table*}
\begin{figure*}[th]
    \centering
    {
\includegraphics[width=\linewidth]{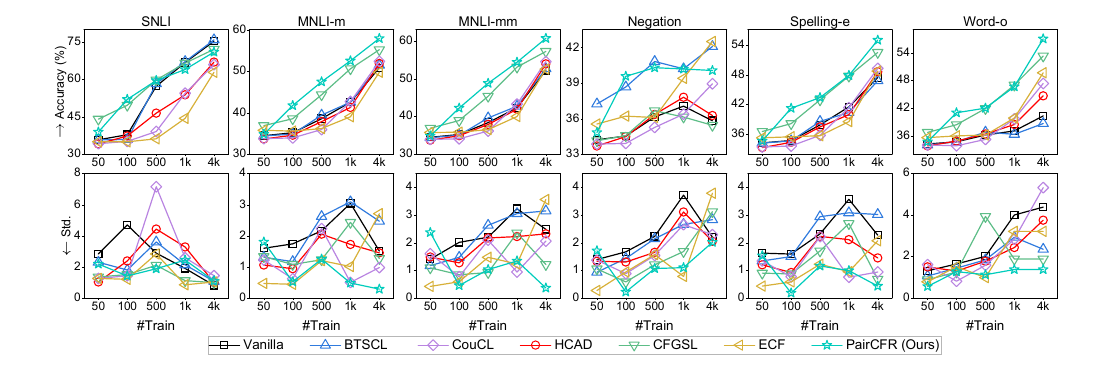}
    }
     \vspace{-0.51cm}
    \caption{Few-shot learning results of BERT$_\mathbf{base}$ on NLI. $x$-axis represents the number of training samples and $y$-axis represents the averaged accuracy and standard deviation on ID and OODs.}
   
    \label{fig:fw_acc_std}
\end{figure*}

\begin{table*}[th]
\caption{The influence of neutral samples during fine-tuning BERT$_\mathbf{base}$ on SNLI. The number of training samples is kept the same. The abbreviations `w' and `w/o' stand for whether neutral examples are included or excluded in the computation of the CL. The p-value is reported under a null hypothesis that no difference exist between training with and without neural samples.}
\label{tab:v2_excluded_n}
\centering
\scalebox{0.8}{
\begin{tabular}{ccccccccc}
\toprule
\multicolumn{1}{c}{\multirow{2}{*}{Train Data}} & \multirow{2}{*}{\begin{tabular}[c]{@{}l@{}}netural\\ samples\end{tabular}} & In-Domain & \multicolumn{6}{c}{Out-of-Dimain} \\ \cmidrule(l){3-9} 
\multicolumn{1}{c}{} &  & SNLI & MNLI-m & MNLI-mm & Negation & Spelling-e & Word-o & $\overline{Acc}$\\ \midrule
PairCAD & w & 73.29±1.09 & 59.41±0.91 & 61.66±0.85 & 35.96±2.81 & 56.42±1.10 & 56.14±2.60 & 53.92 \\
PairCAD & w/o & \textbf{75.38±0.21} & \textbf{60.46±0.38} & \textbf{62.27±0.39} & \textbf{39.21±3.61} & \textbf{56.84±0.54} & \textbf{59.16±0.88} & \textbf{55.59} \\ \midrule
\multicolumn{2}{c}{p-value}& 5.90e-06 & 0.0109 &0.0055&0.0053&0.0087&0.0107&-\\ \bottomrule
\end{tabular}}

\vspace{0.5cm}
\caption{The influence of counterfactual diversity during fine-tuning T5$_\mathbf{base}$ on SNLI. The best results are bolded.}
\label{tab:cad_ratios}
\centering
\tabcolsep=0.2cm
\scalebox{0.8}{
\begin{tabular}{lcccccccc}
\toprule
\multicolumn{1}{c}{} &  & In-Domain & \multicolumn{6}{c}{Out-of-Domain} \\ \cmidrule(l){3-9} 
\multicolumn{1}{c}{\multirow{-2}{*}{\begin{tabular}[c]{@{}c@{}}Train Data\\ CE+CL\end{tabular}}} & \multirow{-2}{*}{R:O} & SNLI & MNLI-m & MNLI-mm & Negation & Spelling-e & Word-o & $\overline{Acc}$ \\ \midrule
Original (20k) & - & \textbf{85.09±0.27} & 69.53±1.38 & 71.62±1.04 & 45.65±3.53 & 66.43±1.49 & 52.89±5.22 & 61.22 \\ \midrule
PairCAD (3.3k) & 1 & 74.50±2.51 & 65.24±1.63 & 67.61±1.36 & 38.38±3.42 & 61.24±1.86 & 60.61±2.33 & 58.62 \\
PairCAD (4.9k) & 2 & 76.12±1.58 & 66.62±1.05 & 69.31±0.87 & 42.33±7.31 & 62.91±1.60 & 62.61±1.58 & 60.76 \\
PairCAD (6.4k) & 3 & {\color[HTML]{000000} 77.98±0.82} & 68.36±1.48 & 70.00±1.44 & 43.13±1.17 & 64.60±1.98 & 64.45±2.15 & 62.11 \\

PairCAD (8.3k) & 4 & 80.14±0.96 & {\color[HTML]{333333} \textbf{71.02±0.39}} & {\color[HTML]{333333} \textbf{71.84±0.76}} & {\color[HTML]{333333} \textbf{45.73±0.70}} & {\color[HTML]{333333} \textbf{66.87±0.51}} & \textbf{67.11±0.39} & \textbf{64.51} \\ \bottomrule

\end{tabular}}
\end{table*}

\section{Results and Analysis}

\subsection{Overall Performance Comparison} 

Table \ref{tab:full} reports the overall performance comparisons, showing that our proposed PairCFR method outperforms all the baseline models on three out of four OOD datasets for both SA and NLI tasks across four different backbone models. To exclude the possibility of marginal improvements due to random initializations, we also conducted significance tests under the null hypothesis that there are no differences between each baseline and our approach, as presented in Table \ref{tab:full_pvalue}, located in Appendix \ref{sec:pvalue}. The p-values less than $0.05$ demonstrate that our methods are significantly better than the baselines, even though some improvements are relatively slight in Table \ref{tab:full}.

In addition, we reported the following findings. Firstly, CAD-based methods may perform worse than non-CAD methods on OOD tasks, e.g., HCAD always lags behind CouCL on the SA task using fine-tuned T5 model. A similar phenomenon is also reported in \cite{joshi-he-2022-investigation}. These could be due to the failure to extract complementary features between CFEs and the original data; Secondly, the introduction of CFEs may shift the training data distribution from the in-domain data distribution. As anticipated, CAD-based methods fall behind non-CAD methods on ID datasets. Thirdly, our proposed PairCFR exhibits superior OOD performance compared to the baselines, achieving the highest accuracy on mostly OOD datasets, with the sole exceptions being the Yelp and Negation datasets. We postulate that the noted exceptions may be attributed to Yelp and Negative datasets having distributions similar to the ID datasets. The above results validate that PairCFR possesses a heightened capability to learn prior knowledge in CAD.


\subsection{Few-shot Learning Performance}
\label{subsec:fewshot}

Data augmentation, such as counterfactual augmentation, is frequently utilized to enhance the performance of few-shot learning. In this part, we investigate the effectiveness of our proposed PairCFR in few-shot learning scenarios.
We conducted experiments using the finetuned BERT$_{base}$ model on the SNLI dataset, gradually increasing the number of training samples from 50 to 4,000. Similarly, on the IMDB dataset, we increased the number of training samples from 32 to 1,024.

Experiment results on SNLI and IMDB under the few-shot setting are reported in Figure \ref{fig:fw_acc_std} and Figure \ref{fig:few-shot imdb} ( Appendix \ref{app:few-shot}). From both tables, we can conclude that our PairCFR generally demonstrates higher accuracy and lower standard deviation across OOD datasets, particularly in scenarios where training sample sizes are small. For instance, PairCFR significantly outperforms other methods by around 6\% on Spelling-e when trained with only 100 counterfactually augmented samples.

\subsection{Ablation Study}
We conducted ablation experiments to verify the efficacy of two crucial strategies of our proposed method: (1) the pairing strategy: the integration of original data with their CFEs within the same batch, denoted \textit{PairCAD}, versus  \textit{ShuffCAD} where randomly shuffle CFEs and originals.
(2) the CL loss: the incorporation of CL and CE loss versus CE loss alone.  

Results in Table \ref{tab:abalation}, together with significance tests in Table \ref{tab:abalation_pvalue} in Appendix \ref{sec:pvalue}, offer several insights: 1) The strategy of pairing original data with their CFEs in the same batch improves OOD performance for both SA and NLI tasks. This can be attributed to the preservation of prior causal relations, which might be lost during random shuffling; 2) The efficacy of PairCAD with a CE-alone learning framework is not guaranteed. For example, within the T5 model framework, PairCAD underperforms ShuffCAD on the SNLI, MNLI, and Spelling-e datasets when only CE loss is adopted. This underscores the critical role of the CL component in augmenting features when we batch CFEs and original data; 
3) Integrating the CL consistently improves model performance in both ID and OODs. Particularly, combining CL with PairCAD yields the best performance across various model assessments, highlighting the effectiveness of contrastive learning and the pairing strategy in leveraging causal relations of CFEs.

\subsection{Impact of Batch Size}
\label{exp:batch_size}
\begin{figure}[tb]
    \flushleft
    \includegraphics[scale=0.415]{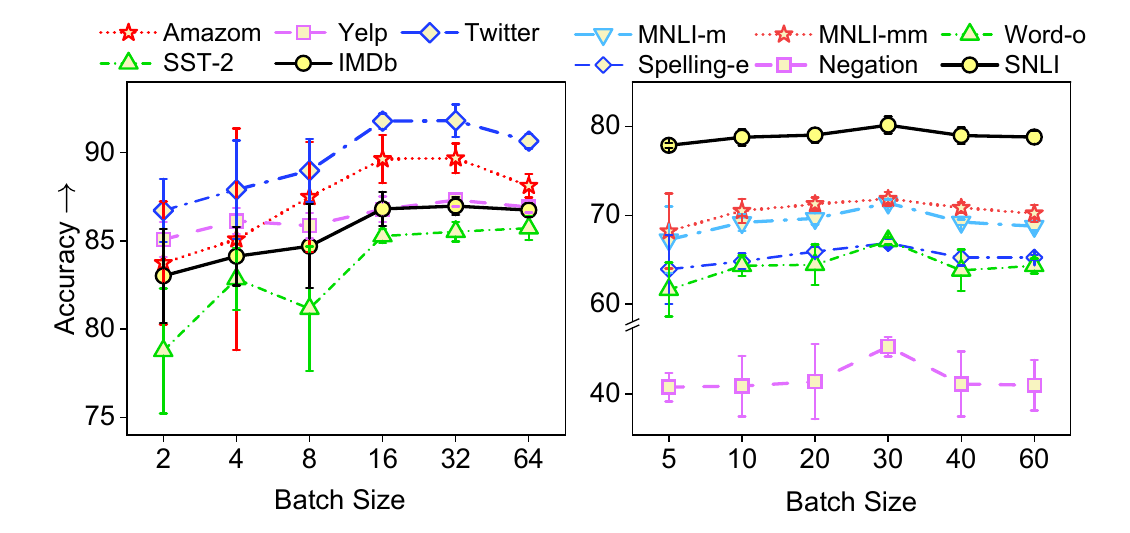}
    \caption{Test results for fine-tuning BERT$_\mathbf{base}$ on IMDb augmented data (left) and T5$_\mathbf{base}$ on SNLI augmented data (right) with respect to the batch size.}
    \label{fig:batchsize}
    \vspace{-0.25cm}
\end{figure}

In this study, we investigated the effect of batch size on learning performance. We conducted experiments on the fine-tuned BERT model for SA and the fine-tuned T5 model for NLI, incrementally increasing the batch size while maintaining the original augmentation ratio for each task. From Figure \ref{fig:batchsize}, we observe that the model performance on both tasks initially improves with increasing batch size, but eventually reaches a plateau or experiences a slight decline. 

We contend that the inclusion of negative samples in the CL function provides additional regularization, forcing the model to rely on a broader array of features beyond those edited. However, an excessively large batch size introduces an overwhelming number of negative samples in CL, which may dilute the human priors in CAD, leading to diminished performance. This trend is consistent across both SA and NLI tasks, highlighting the effort required in batch size selection.

\subsection{Contribution of Neutral Class in NLI}

Do all counterfactual examples equivalently contribute to enhancing model generality? To answer this, we specifically experimented with the fine-tuned BERT model on the NLI task, comparing performance with and without the inclusion of neutral class samples in CL. 

Results in Table \ref{tab:v2_excluded_n} reveal that removing neutral samples, including neutral CFEs, significantly enhances the OOD generalization by approximately 2\% when training the model on CAD with our learning framework. We attribute this performance difference to the distinct nature of neutral samples. In NLI tasks, judgments of entailment and contradiction are often readily determined based on the semantic alignment or disparity between text elements. Conversely, neutral samples represent scenarios where the hypothesis and premise lack any clear relationship, encompassing a vast array of potential expressions. This diversity poses a great challenge for models to identify universal patterns within the neutral class through human annotations. Therefore, adding neutral samples into the CL detrimentally affects the model’s performance in our experiments. 

This investigation highlights the necessity of contemplating the practical value of adding additional counterfactual examples for specific classes.

\subsection{Effect of Counterfactual Diversity}
We also investigated the role of CFE diversity in improving model performance on the NLI task. 
In SNLI, each sentence is annotated with $4$ CFEs, due to the existence of two opposite targets and modifications made to both the hypothesis and premise. Each CFE is obtained through a different type of modification, resulting in a dataset that includes more diverse counterfactuals. 
We fine-tuned the T5$_\mathbf{base}$ model by incrementally including more CFEs in a batch, ranging from $1$ to $4$. 

The results in Table \ref{tab:cad_ratios}, reveal a direct relation between the number of CFEs and the model's generalization capabilities. Notably, the OOD performance of the model trained on CAD is even better than that trained on a $3$ times larger dataset with only original data. We conclude that enhancing counterfactual diversity proves to be an efficient strategy, which is the same as the findings reported in \cite{joshi-he-2022-investigation}. 

%% file: conclusion.tex
\section{Conclusion}
Counterfactually Augmented Data (CAD) can enhance model robustness by explicitly identifying causal features. However, recent research found that CAD may fall behind non-CAD methods on generality. In this work, we introduce PairCFR to overcome this challenge. PairCFR pairs original and counterfactual data during training and includes both contrastive and cross-entropy losses for learning discriminative representations. We prove that contrastive loss aids models in capturing sufficient relationships not represented in CAD, thus improving generality. Extensive experiments demonstrate that our PairCFR achieves superior accuracy and robustness in various scenarios.  Our findings emphasize the potential of carefully designed training paradigms in utilization of CAD.

%% file: limitation.tex
\section{Limitations}


Our PairCFR has been demonstrated to effectively improve models' OOD generalization with human-edited CAD datasets, which, despite its high quality, is quite limited in size. Future work will focus on utilizing LLMs such as ChatGPT or GPT-4 to generate a larger volume of CAD. Yet, LLM-generated CAD may suffer from lower quality due to noisy and insufficient perturbations. It remains crucial and necessary to extend our PairCFR framework to accommodate such compromised CAD. Furthermore, PairCFR currently utilizes a simple form of contrastive loss, namely InfoNCE. In the future, we aim to investigate alternative contrastive loss variants and assess their potential to further enhance OOD generalization capabilities. 
Lastly, our experiments were conducted using relatively older and moderately sized LLMs, such as BERT and RoBERTa. We are also interested in exploring the potential improvements on larger LLMs by employing parameter-efficient finetuning methods.

%% file: ethicsConsideration.tex
\section{Ethics Statement}
This work focuses on reducing shortcut learning in models trained on CAD, thereby improving their robustness and generalization. Similar to other methods designed to mitigate learning from spurious correlations, our proposed PairCFR could help elicit trust in NLP models. It assists models in better-considering context (see Section \ref{sec:method} for details), preventing decision-making based on incomplete or biased information, such as solely on the edited words in CAD. Nonetheless, ensuring absolute fairness in model decisions in complex real-world contexts remains a formidable challenge solely from a model design standpoint. For instance, models could be compromised by low-quality or erroneous counterfactual data, leading to the learning of false relationships and resulting in erroneous or biased real-world decisions. Consequently, it is crucial for practitioners to consider the quality of counterfactual data alongside model design.

\section*{Acknowledgements}
This research is supported, in part, by the Joint NTU-WeBank Research Centre on Fintech, Nanyang Technological University, Singapore. This research is supported, in part, by the National Research Foundation, Prime Minister’s Office, Singapore under its NRF Investigatorship Programme (NRFI Award No. NRF-NRFI05-2019-0002). Any opinions, findings and conclusions or recommendations expressed in this material are those of the author(s) and do not reflect the views of National Research Foundation, Singapore. Xu Guo wants to thank the Wallenberg-NTU Presidential Postdoctoral Fellowship. Zhiwei Zeng thanks the support from the Gopalakrishnan-NTU Presidential Postdoctoral Fellowship. This research is also supported by the Shenzhen Science and Technology Foundation (General Program, JCYJ20210324093212034) and the 2022 Guangdong Province Undergraduate University Quality Engineering Project (Shenzhen University Academic Affairs [2022] No. 7). We also appreciate the support from Guangdong Province Key Laboratory of Popular High Performance Computers 2017B030314073, Guangdong Province Engineering Center of China-made High Performance Data Computing System.


%% file: appendix.tex



\appendix

\clearpage

\section{The trap in the CE loss}
\label{sec:sigmoid_function}
Given a sample, $\mathbf{x}=[x_r, x_c]^{T}$, associated with the label $y\!=\!1$, and the corresponding counterfactual example, $\mathbf{c}=[c_r, x_c]^{T}$, with the flipped label, $y\!=\!0$, by minimizing the cross entropy loss, we compel the model such that $f_W(\mathbf{x})$ approaches $1$ and $f_W(\mathbf{x})$ is close to $0$, respectively. This can be equivalently formulated by maximizing the prediction difference, i.e.,  $\mathrm{max}[{f_W(\mathbf{x})}\!-\!{f_W(\mathbf{c})}]$. The sigmoid function, $\sigma(x)=\frac{1}{1+e^{-x}}$, is \textit{bounded} and \textit{monotonically increasing}, implying that $(w_r x_r + w_c x_c)$ should be as large as possible while $(w_r c_r + w_c x_c)$ should be as small as possible. Here, $x_r$ and $c_r$ are the features before and after editing. The sign of $c_r$ should be opposite to the sign of $x_r$ such that when $f_W(x)$ approaches 1,  $f_W(c)$ can approach 0. For the first term, we observe that increasing $|w_r|$ can lead to an opposite change, i.e., larger $w_r x_r$ and smaller $w_r c_r$. However, the second term, $w_c x_c$, is contained in both $f_W(\mathbf{x})$ and $f_W(\mathbf{c})$. Optimizing $w_c$ does not have the opposite effect. 

\section{Gradient analysis of CL}
\label{sec:gradient_analysis}

In this section, we introduce the details of the gradient of CL with respect to the weight $\mathbf{W}$ through the negative branches $s_{i,n}$.  Before talking details, we rewrite the CL term for convenience, 
\begin{equation}
    \mathcal{L}_{CL} = -\mathop{\mathbb{E}}_{\mathbf{x}_i \in \mathcal{P}_i}\left[\mathrm{log}\frac{e^{s_{ip}/\tau}}{e^{s_{ip}/\tau} + \mathop{\sum}_{\mathbf{x}_n\in\mathcal{N}_i}e^{s_{in}/\tau}}\right].
    \label{eq:clloss9}
\end{equation}

The total derivative of $\mathcal{L}_{CL}$ w.r.t the model weights be calculated through the chain rule as

\begin{align}
 \frac{\partial\mathcal{L}_{CL}}{\partial\mathbf{W}}&=\frac{\partial \mathcal{L}_{CL} }{\partial s_{in}}\times \frac{\partial s_{in}}{\partial \mathbf{W} } + \frac{\partial \mathcal{L}_{CL} }{\partial s_{ip}}\times \frac{\partial s_{ip}}{\partial \mathbf{W} },
\end{align}
where the gradient coming from the branch $s_{in}$ is 
\begin{align}
    \frac{\partial \mathcal{L}_{CL}}{\partial \mathbf{W}}\bigg|_{s_{in}}
    &=\frac{\partial \mathcal{L}_{CL}}{\partial s_{in}}\times \frac{\partial s_{in}}{\partial \mathbf{W}}.
\end{align}
For simplicity, we let $s_{in}=\mathbf{z}_{i}^T\mathbf{z}_n$ and drop the denominator, $\parallel \mathbf{z}_i\parallel\parallel\mathbf{z}_n\parallel$, which is eliminated in the product of partial derivatives. $\mathbf{z}_{i}\!=\!\mathbf{W}^T\mathbf{x}_i$ and $\mathbf{z}_{j}\!=\!\mathbf{W}^T\mathbf{x}_n$, and then we have
\begin{align}
    \frac{\partial s_{in}}{\partial \mathbf{W}} &= \frac{\partial (\mathbf{W}^T\mathbf{x}_i)^T(\mathbf{W}^T\mathbf{x}_n)}{\partial \mathbf{W}}
    \nonumber \\
    &=\frac{\partial (\mathbf{x}_i^T \mathbf{W})(\mathbf{W}^T\mathbf{x}_n)}{\partial \mathbf{W}}
    \nonumber \\
    &=\mathbf{x}_i\mathbf{x}_{n}^T\mathbf{W}+\mathbf{x}_n\mathbf{x}_{i}^T\mathbf{W}
    \nonumber \\
    &=\mathbf{A}_{in}\mathbf{W}
    \label{eqn:gradient_W}.
\end{align}
Here, $\mathbf{A}_{in}\!=\!\mathbf{x}_{i}\mathbf{x}_{n}^{T}+\mathbf{x}_{n}\mathbf{x}_{i}^{T}$. 
The CL term of Eq \eqref{eq:clloss9}
 for anchor $x_i$ can be further written as, 
\begin{small}
    \begin{align}
    \mathcal{L}_{CL}(\mathbf{x}_i) &=-\underset{\mathbf{x}_p\in \mathcal{P}_i}{\mathbb{E}} \left[ \!\log\frac{\exp(s_{ip}/\tau)}{\exp(s_{ip}/\tau)\!+\!\underset{\mathbf{x}_n\in \mathcal{N}_{i}}{\sum}\exp(s_{in}/\tau)}\!\right]
    \nonumber \\
    &= \!\! \underset{\mathbf{x}_p\in \mathcal{P}_i}{\mathbb{E}}\left[ \log \left(\exp(s_{ip}/\tau)\! +  \underset{\mathbf{x}_n\in \mathcal{N}_i}{\sum} \exp(s_{in}/\tau)\right)\right] \nonumber \\
    & \qquad  -\!\underset{\mathbf{x}_p\in \mathcal{P}_i}{\mathbb{E}}\!(s_{ip}/\tau).
\end{align}
\end{small}

Here, only the first term is a function of $s_{i,n}$. Hence, we can compute the gradient of $\mathcal{L}_{CL}$ w.r.t. the similarity for a negative sample, $s_{i,n}$, as follows.
\begin{small}
    \begin{align}
    \frac{\partial \mathcal{L}(\mathbf{x}_i) }{\partial s_{in}}&=\frac{1}{\tau}\underset{\mathbf{x}_p\in \mathcal{P}_i}{\mathbb{E}} \left[\frac{\exp(s_{in}/\tau)}{\exp(s_{ip}/\tau+\underset{\mathbf{x}_n\in \mathcal{N}_{i}}{\sum}\exp(s_{in}/\tau)} \right ]
    \nonumber  \\
    &=\frac{1}{\tau}P_{in} \qquad \text{(written as } P_{in}).
    \label{eq:gradient_sij}
    \end{align}
\end{small}

Combining Eq \eqref{eqn:gradient_W} and Eq \eqref{eq:gradient_sij}  gives the final gradient from a negatives sample,
\begin{align}
\frac{\partial \mathcal{L}(\mathbf{x}_i)}{\partial \mathbf{W}}\bigg|_{s_{in} }
    &=\frac{\partial \mathcal{L}(\mathbf{x}_i)}{\partial s_{in}} \times \frac{\partial s_{in}}{\partial \mathbf{W}}
    \nonumber  \\
    &=\frac{1}{\tau}P_{in} \mathbf{A}_{in} \mathbf{W}. 
    \label{eq:gradient_anegative}
\end{align}

Summing up gradients in Eq \eqref{eq:gradient_anegative} from all negative samples, we can derive
\begin{align}
    \frac{\partial \mathcal{L}(\mathbf{x}_i)}{\partial \mathbf{W}} \bigg|_{\mathcal{N}_i}
    &=\frac{\partial \mathcal{L}(\mathbf{x}_i)}{\partial s_{in}}\times \frac{\partial s_{in}}{\partial \mathbf{W}} \bigg|_{\mathcal{N}_i}
    \nonumber  \\
    &=\frac{1}{\tau} \sum_{\mathbf{x}_n \in \mathcal{N}_i}P_{in} \mathbf{A}_{in} \mathbf{W}.
\end{align}

As the gradient contains pair-wise outer products between the anchor point and all its negative samples, it fully captures the overview of the feature space rather than focusing on a local perspective on edited words.

\section{Experimental Details}

\subsection{Training Data}
\label{apd:train data}
We introduce more details of the CAD data used in model training in our experiments.  We adopt two counterfactually augmented datasets from IMDb \cite{maas2011:imdb} and SNLI \cite{bowman2015snli} in \cite{kaushik2021explaining}. The counterfactually augmented IMDb dataset contains $2440$ original sentences, with each sentence having a corresponding revised counterfactual example. In SNLI, annotators can revise both the hypothesis and the premise for each of two opposite classes, and each sentence has $4$ counterfactual examples. After another round of human filtering, the counterfactual augmented SNLI dataset consists of $9064$ counterfactuals and $2266$ original examples. During training, we split two CAD datasets into train, validation, test sets as shown in Table \ref{tab:traindataset}. 

\begin{table}[h]
    \centering
    \tabcolsep=0.3cm
    \caption{Statistic of human-edited CAD datasets.}
    \label{tab:traindataset}
    \scalebox{0.85}{
    \begin{tabular}{@{}ccccc@{}}
      \hline
        \multicolumn{1}{c}{Dataset} & \multicolumn{1}{c}{\#Train} & \multicolumn{1}{c}{\#Val} & \multicolumn{1}{c}{\#Test} &
        \multicolumn{1}{c}{Total No.}\\ \hline
    \multicolumn{5}{c}{Sentiment Analysis: IMDb} \\ \hline
        Original    & 1707  & 245 & 488 &2440\\
        Revised & 1707  & 245 & 488&2440 \\
        CAD &3414&490&976&4880\\
       \hline
       \multicolumn{5}{c}{Natural Language Inference: SNLI} \\ \hline
        Original & 1666  & 200 & 400 &2266 \\
        Revised & 6664 &800 & 1600 &9064\\
        CAD & 8330  & 1000 & 2000&11330 \\
        \hline
    \end{tabular}}

\vspace{0.5cm}
    \caption{Datasets description. $\sharp$ refers to ID datasets.}
\label{tab:testdataset}
\tabcolsep=0.1cm
\scalebox{0.80}{
\begin{tabular}{@{}llc@{}}
\toprule
\multicolumn{1}{c}{Dataset} & \multicolumn{1}{l}{Domain} & \#Test \\ \midrule
\multicolumn{3}{c}{Sentiment Analysis \#class=2} \\ \midrule
IMDb \cite{maas2011:imdb}$^{\sharp }$ & movie reviews & 67k \\
Amazon \cite{ni2019:amazon} & service feedback & 207k \\
Yelp \cite{zhang2015:yelp} & purchase reviews & 38k \\
Twitter \cite{rosenthal2019:twitter} & social microblogs & 10.3k \\
SST-2 \cite{socher2013:sst-2} & movie reviews & 1.82k \\ \midrule
\multicolumn{3}{c}{Natural Language Inference \#class=3} \\ \midrule
SNLI \cite{bowman2015snli}$^{\sharp }$ & written text & 9.82k \\
MNLI-m \cite{williams2017mnli} & mismatched genres  & 9.83k \\
MNLI-mm \cite{williams2017mnli} & matched genres  & 9.82k \\
Negation \cite{naik2018stress} & strong negation  & 9.83k \\
Spelling-e \cite{naik2018stress} & spelling errors  & 9.14k \\
Word-o \cite{naik2018stress} & large word-overlap  & 9.83k \\ \bottomrule
\end{tabular}}

\end{table}

\subsection{ID and OOD datasets}
\label{apd:id and ood tests}
Here, we provide statistics of in-domain (ID) and out-of-domain (OOD) datasets used to evaluate the generalization of models in Table \ref{tab:testdataset}. 

Since CADs in our experiments are manually revised on samples from IMDb \cite{maas2011:imdb} and SNLI \cite{bowman2015snli}, we include their test datasets for ID evaluation. As for OOD evaluation, we evaluate our sentiment models on Amazon reviews \cite{ni2019:amazon}, Topic-based Tweets sentiment data \cite{rosenthal2019:twitter}, Yelp reviews \cite{zhang2015:yelp} and SST-2 movie reviews \cite{socher2013:sst-2}. On NLI task, we report on the genre-matched (MNLI-m) and genre-mismatched (MNLI-mm) test set of MNLI \cite{williams2017mnli}, which are more challenging than SNLI due to multiple genres. In addition, We additionally employ the diagnostic datasets Negation, Spelling-Error, and Word-Overlap provided by \newcite{naik2018stress} to evaluate models' reasoning abilities on lexical semantics and grammaticality.


\begin{table}[h]
\caption{Model parameter volume in our experiments.}
\label{tab:modelsize}
\centering
\tabcolsep=0.7cm
\scalebox{0.9}{
\begin{tabular}{lc}
    \toprule
        Model & \# Parameters \\ \midrule
        BERT$_{\mathrm{base}}$ & 110M \\
        RoBERTa$_{\mathrm{base}}$ & 125M \\
        SBERT & 250M \\
        T5$_{\mathrm{base}}$ & 223M \\ \bottomrule
\end{tabular}}
\end{table}

\subsection{Implementation details}
\label{apd:hyperparameter values}
In Table \ref{tab:modelsize}, we list the volume of model parameters used in our experiments. In our experiment, we tune hyperparameters of our PairCFR, including learning rate $lr$, batch size $bts$, trade-off factor $\lambda$, and temperature $\tau$, based on the performance on validation set in full dataset finetuning and few shot setting separately. The best hyperparameters are reported in Table \ref{tab:optimal parameters} and Table \ref{tab:few shot parameter}. 


All experiments were conducted on an NVIDIA A100 GPU server equipped with Ubuntu 22.04, featuring 40 GB of GPU memory, 32-core CPUs at 1.5 GHz, and 256 GB of RAM. The test environment was configured with Python 3.8, CUDA 11.8, and Pytorch 2.0. The training time for each hyperparameter configuration is less than one hour.


\begin{figure*}[tb]
    \centering
    \subfloat[{The impact of trade-off term $\lambda$. We fix $\tau\!=\!0.3$ for SA (left) and $\tau\!=\!0.7$ for NLI (right), and gradually increase $\lambda$. }]{\includegraphics[scale=0.43]{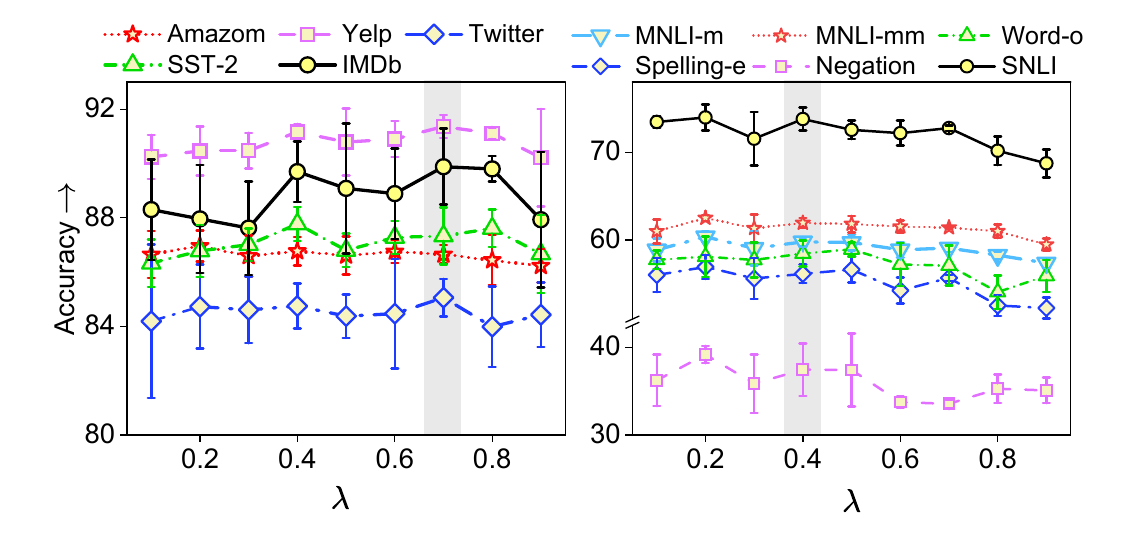}}
    \hspace{0.55cm}
    \subfloat[{The impact of temperature $\tau$. We keep $\lambda=0.7$ for SA (left) and $\lambda = 0.4$ for NLI (right), and gradually increase $\tau$.}]{
    \includegraphics[scale=0.43]{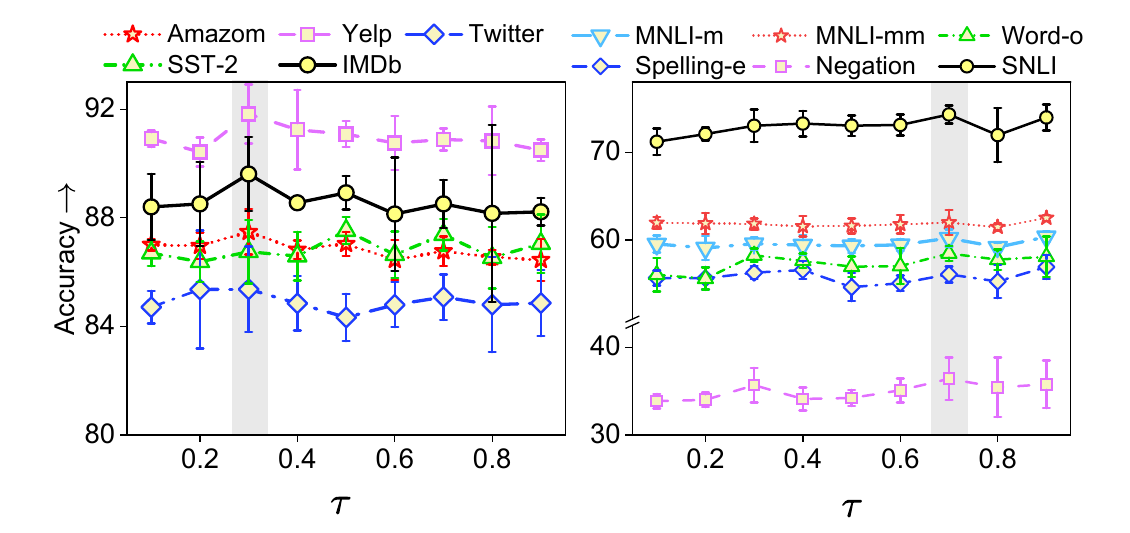}
    }
    
    \caption{The ID and OOD performance of the BERT$_{base}$ models trained on full CAD for IMDb and SNLI tasks. \textcolor[rgb]{0.5,0.5,0.5}{Grey} areas indicate the best hyperparameter settings for $\lambda$ or $\tau$.}
    \label{fig:weight-impact}
\end{figure*}

\begin{figure*}[th]
    \centering
   {\includegraphics[width=0.9\linewidth]{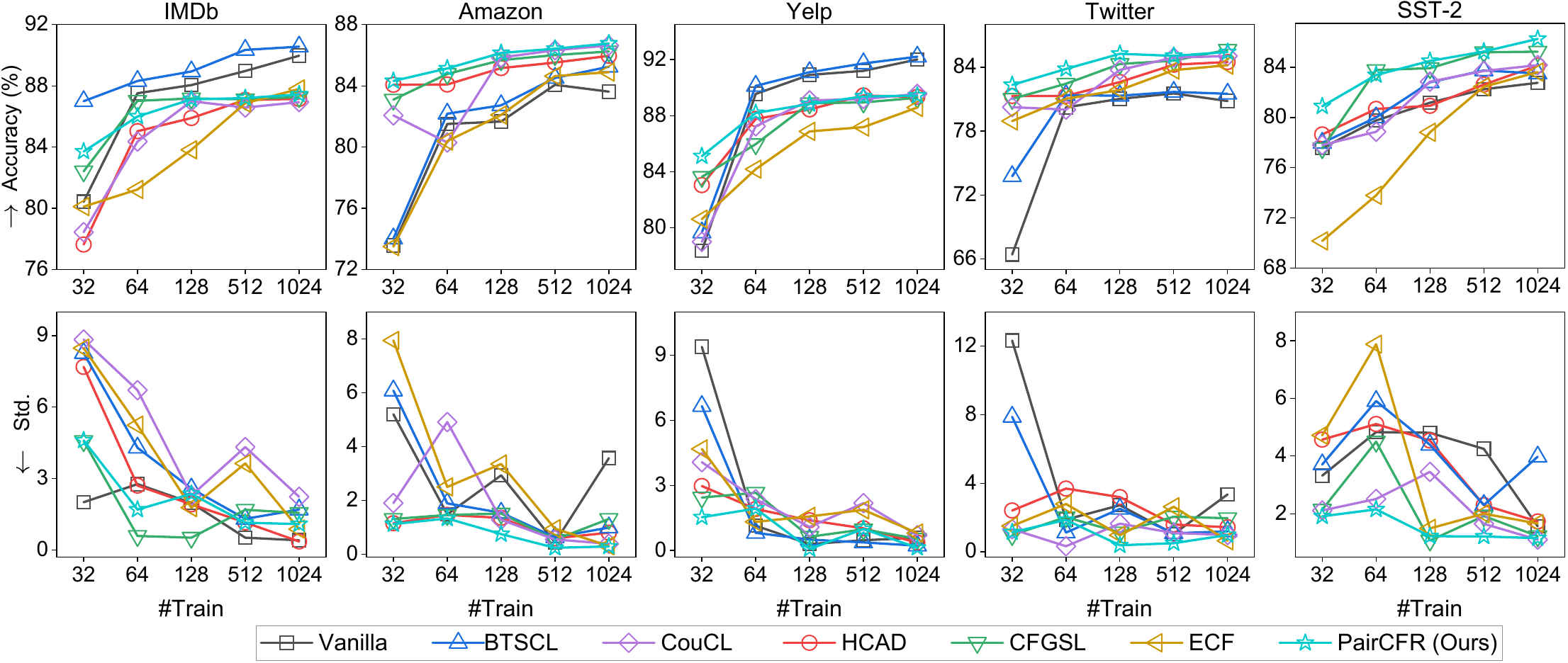}
    }
    \caption{Few-shot learning results of BERT$_\mathbf{base}$ on SA. $x$-axis represents the number of training samples and $y$-axis represents the averaged accuracy and standard deviation on ID and OODs.}
    \label{fig:few-shot imdb}
\end{figure*}


\begin{table*}[t!]
\begin{minipage}[t]{0.47\textwidth}
	\caption{PairCFR hyperparameters for full data finetuning.}
    \label{tab:optimal parameters}
    \centering
    \tabcolsep=0.4cm
    \scalebox{0.85}{
    \begin{tabular}{lcccc}
    \toprule
    \multicolumn{1}{c}{Model} & $lr$ & \multicolumn{1}{l}{$bts$} & $\lambda$ & $\tau$ \\ \midrule
    \multicolumn{5}{c}{Sentiment Analysis} \\  \midrule
    BERT$_\mathbf{base}$ & 3e-5 & 16 & 0.7 & 0.3 \\
    RoBERTa$_\mathbf{base}$ & 3e-6 & 16 & 0.9 & 0.07 \\
    SBERT & 5e-6 & 16 & 0.7 & 0.7 \\
    T5$_\mathbf{base}$ & 1e-4 & 16 & 0.8 & 0.07 \\ \midrule
    \multicolumn{5}{c}{Natural Language Inference} \\  \midrule
    BERT$_\mathbf{base}$ & 3e-5 & 30 & 0.4 & 0.7 \\
    RoBERTa$_\mathbf{base}$ & 1e-5 & 30 & 0.3 & 0.8 \\
    SBERT & 5e-5 & 30 & 0.2 & 0.9 \\
    T5$_\mathbf{base}$ & 1e-4 & 30 & 0.4 & 0.7 \\ \bottomrule
    \end{tabular}}
\end{minipage}
\hspace{0.5cm}
\begin{minipage}[t]{0.47\textwidth}
	\caption{PairCFR hyperparameters for few shot settings. `\#Train' means the training number of shots.}
    \label{tab:few shot parameter}
    \centering
    \tabcolsep=0.32cm
    \scalebox{0.85}{
    \begin{tabular}{lccccc}
    \toprule
    \multicolumn{1}{c}{Model} & \#Train & $lr$ & \multicolumn{1}{l}{$bts$}& $\lambda$& $\tau$ \\ \midrule
    \multicolumn{6}{c}{Sentiment Analysis} \\ \midrule
    \multirow{5}{*}{BERT$_{\mathbf{base}}$} & 32 & 1e-4 & 4 & 0.7&0.3 \\
     & 64 & 1e-5 & 8& 0.7&0.3  \\
     & 128 & 1e-5 & 8  & 0.7&0.3\\
     & 512 & 1e-5 & 16  & 0.7&0.3\\
     & 1024 & 1e-5 & 16 & 0.7&0.3\\ \midrule
    \multicolumn{6}{c}{Natural Language Inference} \\ \midrule
    \multirow{5}{*}{BERT$_{\mathbf{base}}$} & 50 & 1e-5 & 5 & 0.4&0.7  \\
     & 100 & 1e-5 & 5 & 0.4&0.7 \\
     & 500 & 1e-5 & 10 & 0.4&0.7 \\
     & 1k & 1e-5 & 10 & 0.4&0.7 \\
     & 4k & 1e-5 & 20 & 0.4&0.7 \\ \cline{1-6} 
    \end{tabular}}
\end{minipage}
	
\end{table*}

\subsection{Hyperparemeter analysis: $\lambda$ and $\tau$}
In this study, we investigate the influence of trade-off factor $\lambda$ and temperature $\tau$ on model generalization. Specifically, we incrementally increase $\lambda$ or $\tau$ from 0.1 to 0.9 by 0.1 and fix other best hyper-parameters searched from grid search. The experimental results on ID and OODs are reported in Figure \ref{fig:weight-impact}. We observe that with $\lambda$ or $\tau$ increasing from $0.1$, the model performance initially increases and then declines. In SA, the model perform better for a larger $\lambda$ and a lower temperature $0.3$ (i.e., $\lambda\!=\!0.7,\tau\!=\!0.3$), while in NLI, a larger temperature and smaller $\lambda$ is favored (i.e., $\lambda\!=\!0.4,\tau\!=\!0.7$). We hypothesize that in SA, the model may overly depend on perturbed words for predictions, as revision patterns are relatively smaller than in NLI. Therefore, we should incorporate a smaller temperature $\tau$ and a higher trade-off $\lambda$ to introduce a higher regularization from contrastive learning in SA. More insights will be explored in future work.



\subsection{Few-shot learning on SA}
\label{app:few-shot}

Here, we present the results of few-shot learning using the BERT model on the SA task, with the number of IMDb augmented data progressively increasing from 32 to 1024, as shown in Figure \ref{fig:few-shot imdb}. Similar to the trend observed in few-shot learning for the NLI task, discussed in Section \ref{subsec:fewshot}, our approach demonstrates significant performance improvements even with limited data in the SA task.

\begin{table*}[]
\caption{{Results of statistical significance test under the hypothesis that there are no differences between baselines and our approach on both ID and OOD. P-values less than 0.05 are bolded, indicating a substantive disparity between two methods.}}
\label{tab:full_pvalue}
\tabcolsep=0.15cm
\scalebox{0.75}{
\begin{tabular}{@{}lccccccccccc@{}}
\toprule
\multicolumn{6}{c}{Sentiment Analysis} & \multicolumn{6}{c}{Natural Language Inference} \\ \midrule
\multicolumn{1}{c}{} & \multicolumn{1}{c|}{In-Domain} & \multicolumn{4}{c||}{Out-of-Dimain} & \multicolumn{1}{c|}{In-Domain} & \multicolumn{5}{c}{Out-of-Dimain} \\ \cmidrule(l){2-12} 
\multicolumn{1}{c}{\multirow{-2}{*}{\begin{tabular}[c]{@{}c@{}}Baseline \\ vs. Ours\end{tabular}}} & \multicolumn{1}{c|}{IMDb} & Amazon & Yelp & Twitter & \multicolumn{1}{c||}{SST-2} & \multicolumn{1}{c|}{SNLI} & MNLI-m & MNLI-mm & Negation & Spelling-e & Word-o \\ \midrule
\multicolumn{12}{c}{BERT-base-uncased} \\ \midrule
Vanilla & \multicolumn{1}{c|}{0.3237} & \textbf{0.0495} & \textbf{0.0043} & \textbf{2.40E-06} & \multicolumn{1}{c||}{\textbf{1.63E-05}} & \multicolumn{1}{c|}{\textbf{0.0012}} & \textbf{0.0136} & \textbf{0.0111} & 0.5754 & \textbf{0.0140} & 0.1182 \\
BTSCL & \multicolumn{1}{c|}{0.0665} & \textbf{0.0411} & 0.1075 & \textbf{0.0005} & \multicolumn{1}{c||}{\textbf{2.11E-06}} & \multicolumn{1}{c|}{\textbf{2.75E-05}} & \textbf{0.0044} & \textbf{0.0053} & 0.1491 & \textbf{0.0101} & \textbf{0.0047} \\
CouCL & \multicolumn{1}{c|}{\textbf{7.72E-06}} & 0.8357 & \textbf{6.10E-05} & \textbf{0.0204} & \multicolumn{1}{c||}{\textbf{0.0012}} & \multicolumn{1}{c|}{\textbf{0.0005}} & \textbf{0.0001} & \textbf{8.17E-05} & 0.7220 & \textbf{0.0007} & \textbf{0.0004} \\
HCAD & \multicolumn{1}{c|}{0.077} & 0.1308 & \textbf{0.001} & \textbf{0.0498} & \multicolumn{1}{c||}{\textbf{0.002}} & \multicolumn{1}{c|}{\textbf{0.0323}} & \textbf{0.0382} & 0.0637 & 0.1826 & 0.0590 & 0.0588 \\
CFGSL & \multicolumn{1}{c|}{0.3011} & \textbf{0.0457} & \textbf{0.0141} & \textbf{0.0421} & \multicolumn{1}{c||}{0.5235} & \multicolumn{1}{c|}{\textbf{1.06E-05}} & 0.0932 & 0.8232 & \textbf{0.0018} & \textbf{0.0078} & \textbf{0.0040} \\
ECF & \multicolumn{1}{c|}{\textbf{0.0279}} & \textbf{0.0457} & \textbf{6.61E-06} & \textbf{0.0003} & \multicolumn{1}{c||}{0.2573} & \multicolumn{1}{c|}{0.1848} & \textbf{0.0177} & 0.0867 & 0.3704 & \textbf{0.0346} & 0.1361 \\ \midrule
\multicolumn{12}{c}{RoBERTa-base} \\ \midrule
Vanilla & \multicolumn{1}{c|}{\textbf{0.0448}} & \textbf{0.046} & \textbf{0.0495} & \textbf{0.0029} & \multicolumn{1}{c||}{\textbf{0.0469}} & \multicolumn{1}{c|}{\textbf{1.45E-06}} & \textbf{0.0102} & 0.0715 & 0.2057 & \textbf{0.0225} & \textbf{0.0452} \\
BTSCL & \multicolumn{1}{c|}{\textbf{0.0394}} & 0.2731 & \textbf{0.0019} & \textbf{0.0231} & \multicolumn{1}{c||}{\textbf{0.0266}} & \multicolumn{1}{c|}{\textbf{6.26E-05}} & \textbf{0.0484} & 0.3835 & 0.9955 & \textbf{0.0344} & \textbf{0.0076} \\
CouCL & \multicolumn{1}{c|}{\textbf{0.0410}} & 0.1456 & \textbf{0.0462} & \textbf{0.0443} & \multicolumn{1}{c||}{\textbf{0.0182}} & \multicolumn{1}{c|}{0.0922} & 0.0584 & \textbf{0.0207} & \textbf{0.0396} & 0.1400 & \textbf{0.0376} \\
HCAD & \multicolumn{1}{c|}{\textbf{0.0442}} & \textbf{0.0349} & \textbf{0.0154} & \textbf{0.0014} & \multicolumn{1}{c||}{\textbf{0.0029}} & \multicolumn{1}{c|}{\textbf{6.51E-05}} & \textbf{0.0030} & \textbf{0.0005} & \textbf{0.0008} & \textbf{0.0180} & \textbf{0.0286} \\
CFGSL & \multicolumn{1}{c|}{\textbf{0.0317}} & \textbf{0.0241} & 0.1834 & \textbf{0.0007} & \multicolumn{1}{c||}{\textbf{0.0380}} & \multicolumn{1}{c|}{\textbf{0.0348}} & 0.3550 & \textbf{0.0496} & \textbf{0.0033} & 0.7874 & \textbf{0.0014} \\
ECF & \multicolumn{1}{c|}{\textbf{0.0361}} & \textbf{0.031} & \textbf{0.0012} & \textbf{0.0012} & \multicolumn{1}{c||}{\textbf{0.0021}} & \multicolumn{1}{c|}{\textbf{0.0167}} & \textbf{0.0147} & \textbf{0.0112} & 0.0830 & \textbf{0.0121} & \textbf{0.0071} \\ \midrule
\multicolumn{12}{c}{SBERT-multi-qa-cos} \\ \midrule
Vanilla & \multicolumn{1}{c|}{0.4796} & \textbf{1.56E-08} & \textbf{0.0002} & \textbf{3.66E-05} & \multicolumn{1}{c||}{\textbf{0.0003}} & \multicolumn{1}{c|}{\textbf{6.48E-05}} & \textbf{0.0273} & \textbf{0.0132} & \textbf{0.0076} & \textbf{0.0383} & \textbf{0.0306} \\
BTSCL & \multicolumn{1}{c|}{\textbf{0.0470}} & \textbf{1.71E-07} & \textbf{2.01E-11} & \textbf{1.11E-07} & \multicolumn{1}{c||}{\textbf{4.70E-03}} & \multicolumn{1}{c|}{\textbf{2.94E-05}} & \textbf{0.0138} & \textbf{0.003} & \textbf{0.0006} & \textbf{0.04611} & \textbf{0.0035} \\
CouCL & \multicolumn{1}{c|}{\textbf{0.0097}} & \textbf{0.0001} & \textbf{0.0002} & \textbf{0.0004} & \multicolumn{1}{c||}{\textbf{0.0099}} & \multicolumn{1}{c|}{\textbf{0.0403}} & \textbf{0.0448} & \textbf{0.0275} & 0.1397 & 0.0569 & \textbf{0.0428} \\
HCAD & \multicolumn{1}{c|}{\textbf{0.0173}} & \textbf{7.43E-05} & \textbf{0.0025} & \textbf{0.0221} & \multicolumn{1}{c||}{\textbf{4.51E-05}} & \multicolumn{1}{c|}{\textbf{0.0051}} & 0.0584 & \textbf{0.0457} & 0.079 & \textbf{0.0422} & \textbf{0.0498} \\
CFGSL & \multicolumn{1}{c|}{\textbf{0.0050}} & \textbf{0.0006} & \textbf{0.008} & \textbf{4.22E-06} & \multicolumn{1}{c||}{0.0197} & \multicolumn{1}{c|}{\textbf{0.0325}} & \textbf{0.0421} & \textbf{0.03106} & 0.485 & \textbf{0.0215} & {\color[HTML]{333333} 0.0533} \\
ECF & \multicolumn{1}{c|}{0.0959} & 0.4188 & 0.3184 & \textbf{0.0013} & \multicolumn{1}{c||}{0.3667} & \multicolumn{1}{c|}{\textbf{0.0008}} & \textbf{0.0019} & \textbf{0.0013} & 0.1876 & \textbf{0.0019} & \textbf{0.0017} \\ \midrule
\multicolumn{12}{c}{T5-base} \\ \midrule
Vanilla & \multicolumn{1}{c|}{0.1072} & \textbf{0.0144} & \textbf{6.37E-05} & \textbf{0.0002} & \multicolumn{1}{c||}{\textbf{0.0112}} & \multicolumn{1}{c|}{\textbf{0.0216}} & \textbf{0.0299} & \textbf{0.0162} & \textbf{0.0294} & \textbf{0.0302} & \textbf{0.0088} \\
BTSCL & \multicolumn{1}{c|}{\textbf{0.0025}} & \textbf{0.0300} & \textbf{8.42E-05} & \textbf{8.42E-05} & \multicolumn{1}{c||}{\textbf{8.42E-05}} & \multicolumn{1}{c|}{\textbf{0.0207}} & \textbf{0.0468} & \textbf{0.0356} & \textbf{0.0349} & \textbf{0.0445} & \textbf{0.0411} \\
CouCL & \multicolumn{1}{c|}{\textbf{0.0464}} & 0.1554 & \textbf{0.03123} & \textbf{0.019} & \multicolumn{1}{c||}{\textbf{0.0027}} & \multicolumn{1}{c|}{\textbf{0.0319}} & \textbf{0.0407} & \textbf{0.0459} & \textbf{0.0397} & \textbf{0.0309} & \textbf{0.0211} \\
HCAD & \multicolumn{1}{c|}{\textbf{0.0306}} & 0.1720 & \textbf{0.0012} & \textbf{0.0028} & \multicolumn{1}{c||}{\textbf{0.0001}} & \multicolumn{1}{c|}{\textbf{0.0463}} & 0.0566 & 0.0772 & 0.4857 & \textbf{0.0421} & \textbf{0.0438} \\
CFGSL & \multicolumn{1}{c|}{0.4158} & 0.1139 & 0.2299 & \textbf{0.0067} & \multicolumn{1}{c||}{\textbf{0.0014}} & \multicolumn{1}{c|}{\textbf{0.0497}} & \textbf{0.0452} & \textbf{0.0229} & 0.4665 & \textbf{0.0416} & {\color[HTML]{333333} 0.0721} \\
ECF & \multicolumn{1}{c|}{\textbf{0.0053}} & 0.2914 & \textbf{0.0065} & 0.1045 & \multicolumn{1}{c||}{0.4612} & \multicolumn{1}{c|}{\textbf{0.0352}} & 0.0976 & \textbf{0.0452} & 0.4403 & \textbf{0.0321} & 0.0813 \\ \bottomrule
\end{tabular}}


\vspace{0.5cm}
\caption{Results of statistical significance test under the hypothesis that there are no differences between two ablation studies. P-values less than 0.05 are bolded, indicating a substantive disparity.}
\label{tab:abalation_pvalue}
\tabcolsep=0.11cm
\scalebox{0.7}{
\begin{tabular}{@{}llccccccccccc@{}}
\toprule
\multicolumn{2}{c}{} & \multicolumn{5}{c}{Sentiment Analysis} & \multicolumn{6}{c}{Natural Language Inference} \\ \midrule
\multicolumn{2}{c}{Variants} & \multicolumn{1}{c|}{In-Domain} & \multicolumn{4}{c||}{Out-of-Dimain} & \multicolumn{1}{c|}{In-Domain} & \multicolumn{5}{c}{Out-of-Dimain} \\ \midrule
Control & \multicolumn{1}{c}{Comparison} & \multicolumn{1}{c|}{IMDb} & Amazon & Yelp & Twitter & \multicolumn{1}{c||}{SST-2} & \multicolumn{1}{c|}{SNLI} & MNLI-m & MNLI-mm & Negation & Spelling-e & Word-o \\ \midrule
\multicolumn{13}{c}{BERT-base-uncased} \\ \midrule
CE & Shuff vs. Pair & \multicolumn{1}{c|}{0.8727} & 0.5053 & 0.9418 & 0.3465 & \multicolumn{1}{c||}{0.4981} & \multicolumn{1}{c|}{0.1934} & 0.2881 & 0.7977 & 0.4542 & \textbf{0.0450} & 0.3317 \\
CE+CL & Shuff vs. Pair & \multicolumn{1}{c|}{\textbf{0.0389}} & 0.9057 & \textbf{0.0055} & 0.1469 & \multicolumn{1}{c||}{\textbf{0.0350}} & \multicolumn{1}{c|}{\textbf{0.0120}} & \textbf{0.0011} & 0.1890 & \textbf{0.0268} & \textbf{0.0008} & \textbf{0.0406} \\
ShuffCAD & CE vs. CE+CL & \multicolumn{1}{c|}{0.2311} & 0.1238 & \textbf{0.0018} & 0.1890 & \multicolumn{1}{c||}{\textbf{0.0155}} & \multicolumn{1}{c|}{0.5306} & 0.1866 & 0.0973 & 0.2621 & 0.1722 & 0.0736 \\
PairCAD & CE vs. CE+CL & \multicolumn{1}{c|}{0.2021} & 0.3666 & \textbf{0.0417} & \textbf{0.0395} & \multicolumn{1}{c||}{\textbf{0.0032}} & \multicolumn{1}{c|}{\textbf{0.0135}} & \textbf{0.0034} & \textbf{0.0137} & \textbf{0.0293} & 0.2280 & \textbf{0.0210} \\ \midrule
\multicolumn{13}{c}{RoBERTa-base} \\ \midrule
CE & Shuff vs. Pair & \multicolumn{1}{c|}{\textbf{0.0376}} & \textbf{0.0045} & \textbf{0.0049} & 0.9751 & \multicolumn{1}{c||}{0.4894} & \multicolumn{1}{c|}{\textbf{0.0246}} & 0.3194 & {0.0723} & \textbf{0.0031} & 0.2519 & 0.3029 \\
CE+CL & Shuff vs. Pair & \multicolumn{1}{c|}{0.3722} & 0.1181 & 0.0720 & 0.3250 & \multicolumn{1}{c||}{\textbf{0.0009}} & \multicolumn{1}{c|}{0.2123} & \textbf{0.0037} & 0.3623 & \textbf{0.0072} & \textbf{0.0033} & \textbf{0.0007} \\
ShuffCAD & CE vs. CE+CL & \multicolumn{1}{c|}{\textbf{0.0005}} & \textbf{2.48E-06} & \textbf{6.52E-05} & \textbf{8.76E-07} & \multicolumn{1}{c||}{\textbf{0.0073}} & \multicolumn{1}{c|}{\textbf{0.0004}} & \textbf{0.0178} & \textbf{0.0016} & \textbf{0.0006} & 0.0540 & {0.0655} \\
PairCAD & CE vs. CE+CL & \multicolumn{1}{c|}{\textbf{0.0298}} & \textbf{0.0133} & 0.1017 & \textbf{0.0120} & \multicolumn{1}{c||}{\textbf{0.0011}} & \multicolumn{1}{c|}{0.1585} & \textbf{0.0012} & \textbf{0.0252} & 0.2565 & \textbf{0.0040} & 0.2420 \\ \midrule
\multicolumn{13}{c}{SBERT-multi-qa-distilbert-cos} \\ \midrule
CE & Shuff vs. Pair & \multicolumn{1}{c|}{\textbf{0.0058}} & \textbf{9.65E-09} & \textbf{0.0011} & 0.6263 & \multicolumn{1}{c||}{\textbf{0.0086}} & \multicolumn{1}{c|}{\textbf{0.0491}} & 0.3697 & 0.3337 & 0.2248 & 0.5029 & 0.4971 \\
CE+CL & Shuff vs. Pair & \multicolumn{1}{c|}{0.1317} & \textbf{0.0004} & 0.4958 & \textbf{1.65E-07} & \multicolumn{1}{c||}{\textbf{0.0027}} & \multicolumn{1}{c|}{0.6576} & 0.6699 & \textbf{0.0187} & \textbf{0.0476} & 0.4170 & 0.1311 \\
ShuffCAD & CE vs. CE+CL & \multicolumn{1}{c|}{\textbf{0.0003}} & \textbf{4.29E-04} & \textbf{3.43E-06} & \textbf{0.0049} & \multicolumn{1}{c||}{\textbf{0.0021}} & \multicolumn{1}{c|}{0.4285} & 0.0930 & 0.1230 & 0.5577 & \textbf{0.0494} & 0.1786 \\
PairCAD & CE vs. CE+CL & \multicolumn{1}{c|}{0.1491} & \textbf{1.41E-06} & 0.6202 & \textbf{0.0002} & \multicolumn{1}{c||}{\textbf{0.0002}} & \multicolumn{1}{c|}{0.2408} & 0.1698 & \textbf{0.0011} & 0.1113 & \textbf{0.0335} & \textbf{0.0118} \\ \midrule
\multicolumn{13}{c}{T5-base} \\ \midrule
CE & Shuff vs. Pair & \multicolumn{1}{c|}{0.8304} & 0.1841 & 0.1112 & \textbf{0.0013} & \multicolumn{1}{c||}{\textbf{0.0006}} & \multicolumn{1}{c|}{\textbf{0.0024}} & \textbf{2.99E-06} & \textbf{0.0006} & 0.9644 & \textbf{0.0003} & \textbf{0.0002} \\
CE+CL & Shuff vs. Pair & \multicolumn{1}{c|}{\textbf{0.0029}} & 0.1851 & \textbf{0.0096} & \textbf{0.0108} & \multicolumn{1}{c||}{\textbf{0.0004}} & \multicolumn{1}{c|}{0.2966} & \textbf{0.0042} & \textbf{0.0415} & 0.5030 & 0.1371 & 0.1530 \\
ShuffCAD & CE vs. CE+CL & \multicolumn{1}{c|}{0.4340} & 0.1206 & 0.0876 & 0.4625 & \multicolumn{1}{c||}{\textbf{0.0164}} & \multicolumn{1}{c|}{0.5484} & 0.4942 & 0.4354 & 0.4859 & 0.4489 & 0.2817 \\
PairCAD & CE vs. CE+CL & \multicolumn{1}{c|}{\textbf{0.0029}} & 0.1837 & \textbf{0.0096} & \textbf{0.0108} & \multicolumn{1}{c||}{\textbf{0.0004}} & \multicolumn{1}{c|}{\textbf{0.0481}} & \textbf{0.0098} & \textbf{0.0223} & 0.4851 & \textbf{0.0284} & \textbf{0.0497} \\ \bottomrule
\end{tabular}}
\end{table*}

\subsection{Statistical significance test}
\label{sec:pvalue}

To ensure that the observed improvements are not due to randomness across multiple trials, we conducted statistical significance tests on comparative experiments and ablation studies. We first check that experimental results from random initialization on both ID and OOD datasets follow a Gaussian distribution, and thus employ a two-sided paired samples T-test. Our T-tests are conducted under the null hypothesis that there are no differences between the two groups of experiments.

Table \ref{tab:full_pvalue} presents the significance test results of our method against all baselines for the comparative experiments (refer to Table \ref{tab:full}). We observed that the majority of p-values fall below the conventional confidence level of 0.05, indicating that the improvements in OOD performance achieved by our algorithm over the baselines are statistically significant and not due to randomness. Similarly, Table \ref{tab:abalation_pvalue} presents the significance test results of the ablation study (refer to Table \ref{tab:abalation}), verifying the effectiveness of our pairing strategy and CL function.
